\documentclass[11pt]{article}

\usepackage[preprint]{acl}

\usepackage{times}
\usepackage{latexsym}

\usepackage[T1]{fontenc}

\usepackage[utf8]{inputenc}

\usepackage{microtype}

\usepackage{inconsolata}

\usepackage{graphicx}

\usepackage{algorithm}
\usepackage{amsmath}
\usepackage{amssymb}
\usepackage{makecell}
\usepackage{booktabs}       
\usepackage{amsfonts}       
\usepackage{nicefrac}       
\usepackage{xcolor}         
\usepackage{varwidth}
\usepackage{subcaption}
\usepackage{adjustbox}
\usepackage{colortbl}
\usepackage{multirow}
\usepackage{multicol}
\usepackage{algpseudocode}
\usepackage{amsthm}
\newtheorem{lemma}{Lemma}
\definecolor{mygreen}{RGB}{0,150,0}
\definecolor{myred}{RGB}{150,0,0}

\usepackage{newfloat}
\usepackage{listings}

\title{{\textbf{\textit{A}}dvantageous \textbf{\textit{P}}arameter \textbf{\textit{EX}}pansion Training \\ Makes Better Large Language Models}}

\author{Naibin Gu\textsuperscript{\rm 1,2}\thanks{\ \  Equal contribution. $^\dagger$ Corresponding author. $^\ddagger$ Project lead.},\ Yilong Chen\textsuperscript{\rm 1,2}\footnotemark[1],\ Zhenyu Zhang\textsuperscript{\rm 3}$^{\ddagger}$,\ Peng Fu\textsuperscript{\rm 1,2}$^{\dagger}$,\ Zheng Lin\textsuperscript{\rm 1,2},\\ {\bf Shuohuan Wang\textsuperscript{\rm 3},\ Yu Sun\textsuperscript{\rm 3},\ Hua Wu\textsuperscript{\rm 3},\ Weiping Wang\textsuperscript{\rm 1},\ Haifeng Wang\textsuperscript{\rm 3}} \\ 
\textsuperscript{\rm 1}Institute of Information Engineering, Chinese Academy of Sciences, Beijing, China \\
\textsuperscript{\rm 2}School of Cyber Security, University of Chinese Academy of Sciences, Beijing, China \\
\textsuperscript{\rm 3}Baidu Inc., Beijing, China \\
  \texttt{ \textrm{\{}gunaibin,chenyilong,fupeng\textrm{\}}@iie.ac.cn} \\
  \texttt{ \textrm{\{}zhangzhenyu07,wangshuohuan\textrm{\}}@baidu.com} \\
}

\begin{document}
\maketitle
\begin{abstract}
Although scaling up the number of trainable parameters can effectively improve the training performance of large language models, it also leads to increased computational overhead. When delving into the parameter difference, we find that a subset of parameters, termed advantageous parameters, plays a crucial role in determining model performance. Further analysis reveals that stronger models tend to possess more such parameters. In this paper, we propose \textbf{A}dvantageous \textbf{P}arameter \textbf{EX}pansion Training (APEX), a method that progressively expands advantageous parameters into the space of disadvantageous ones, thereby increasing their proportion and enhancing training effectiveness, while keeping the total parameter count unchanged. Extensive experiments on both instruction tuning and continued pre-training across five base models demonstrate that, in instruction tuning, APEX outperforms full-parameter tuning while using only 52\% of the trainable parameters. In continued pre-training, APEX achieves the same perplexity level as conventional training with only approximately 30\% of the training data, and yields significant improvements on downstream tasks\footnote{The code is available at \url{https://github.com/gccnlp/APEX}}.
\end{abstract}

\section{Introduction}
Large language models (LLMs) have demonstrated remarkable performance across various natural language processing tasks~\cite{openai2024gpt4technicalreport,Touvron2023LLaMAOA,touvron2023llama2openfoundation,wang2026ernie50technicalreport}. 
A prevailing strategy to enhance the model's performance is to scale up trainable parameters, as evidenced in both pre-training~\cite{deepseekai2025deepseekv3technicalreport,cai2024internlm2technicalreport,yang2024qwen2technicalreport} and fine-tuning~\cite{ivison2023camelschangingclimateenhancing,meng2024pissa}: 
more parameters tend to improve effectiveness.
However, despite its success, this scaling paradigm entails substantial computational cost, posing significant challenges for both training and deployment.

Intriguingly, large language models are known to exhibit substantial parameter redundancy~\cite{DBLP:conf/iclr/FrankleC19,ma2023llmpruner,gu-etal-2024-light}. Recent studies have revealed that this redundancy is highly structured, with parameters associated with 
high-magnitude activations disproportionately controlling information 
flow and model outputs~\cite{muralidharan2024compactlanguagemodelspruning,
lin2024awqactivationawareweightquantization,DBLP:conf/aaai/AnZYTW24}. 
For example, removing up to 50\% of parameters with low activation 
magnitudes causes negligible performance loss~\cite{sun2024a}, 
suggesting that the model capability concentrates in a subset of 
parameters. 
Herein, we refer to these influential parameters as \textbf{advantageous parameters}, while noting that a substantial portion of capacity remains underutilized. 
This observation raises a pivotal research question:

\textit{Can we optimize training strategies to increase the proportion of advantageous parameters, thereby maximizing model capabilities without merely scaling up the parameters?}

To answer this question, we first analyze activation distributions across 
different components in modern LLMs, confirming that high-activation 
parameters consistently dominate model outputs. On this basis, we investigate the relationship between 
activation distribution and model capability. By comparing several groups of strong and weak models (i.e., higher vs. lower benchmark performance) of the same scale, we observe that weak models tend to exhibit sharper activation distributions, suggesting that model output is dominated by a subset of parameters while the majority remain underutilized. In contrast, strong models display flatter, more balanced distributions, indicating a broader set of advantageous parameters and more uniform contributions. Based on this observation, we hypothesize that model performance is not solely dictated by the total parameter budget, but by how effectively that budget is utilized.

In this paper, we present \textbf{A}dvantageous \textbf{P}arameter \textbf{EX}pansion Training (APEX), a novel training strategy that expands the proportion of advantageous parameters without altering the model scale and architecture.
Specifically, to progressively expand the advantageous parameter space, 
APEX divides training into multiple stages. At the beginning of each stage, APEX selects the sets of advantageous and disadvantageous parameters based on the assessment results from the previous stage. The assessment is performed by tracking the activations of heads in the Multi-Head Attention (MHA) and channels in the Feed-Forward Network (FFN) during each forward pass. Building upon the assessment, APEX introduces expansion operators into each MHA and FFN module. In MHA, the head expansion operators expand the parameter space from advantageous heads to disadvantageous ones, and similarly for channels in the FFN, thereby increasing the proportion of advantageous parameters. By initializing these operators to zero at the start and fusing them back into the model at the end of each stage, APEX preserves the original architecture and integrates smoothly into standard training procedures.

We conduct extensive experiments in two common training scenarios: instruction tuning and continued pre-training, using five base models of different sizes ranging from 1B to 13B parameters. In the instruction tuning scenario, experiments show that APEX consistently outperforms 9 strong baselines, including parameter-efficient tuning, half-parameter tuning, and full-parameter tuning methods. Remarkably, it achieves better performance than full-parameter tuning using only 52\% of the trainable parameters. In the continued pre-training scenario, APEX achieves the perplexity level of conventional training using only approximately 30\% of the data budget, demonstrating a convergence speedup. Furthermore, evaluations on ten representative downstream tasks demonstrate that APEX significantly outperforms conventional training. These empirical results are further supported by both observations of activation distribution shifts and a simplified theoretical model (Appendix~\ref{method-theory}), indicating that APEX improves the utilization of previously under-used subspaces. Overall, APEX demonstrates a strong capability to enhance model training performance through advantageous parameter expansion.
\section{Observations}
\label{sec-obs}
\noindent \textbf{Background and Notations.}
In Transformer-based models, each layer mainly consists of a Multi-Head Attention (MHA) module and a Feed-Forward Network (FFN) module. Each MHA module comprises $H$ attention heads parameterized by projection matrices $\mathbf{W}_{Q}^{h}$, $\mathbf{W}_{K}^{h}$, $\mathbf{W}_{V}^{h} \in \mathbb{R}^{d_\text{model}\times d_\text{head}}$ and $\mathbf{W}_{O} \in \mathbb{R}^{d_{\text{model}}\times d_\text{model}}$, where $d_\text{head}=d_\text{model}/H$. 
Given an input $\mathbf{X}$, we have:
\begin{equation}
\begin{aligned}
    \text{MHA}(\mathbf{X})={\rm Concat}(\text{head}_{1},..., \text{head}_{H})\mathbf{W}_O,
\end{aligned}
\end{equation}

Most LLMs employ a Gated Linear Units structure \cite{DBLP:conf/icml/DauphinFAG17} in FFNs, and each module involves three weight matrices: $\mathbf{W}_{U}, \mathbf{W}_{G} \in \mathbb{R}^{d_\text{model}\times d_\text{ffn}}$ for upward projection and gating, and $\mathbf{W}_{D} \in \mathbb{R}^{d_\text{ffn}\times d_\text{model}}$. For the input $\mathbf{X}$, we have:
\begin{equation}
\begin{aligned}
    \text{FFN}(\mathbf{X})=((\mathbf{X}\mathbf{W}_U)\odot f(\mathbf{X}\mathbf{W}_G))\mathbf{W}_D,
\end{aligned}
\end{equation}
where $f(\cdot)$ denotes the activation function and $\odot$ denotes the element-wise multiplication. 
\begin{figure}[t]
  \subfloat[MHA Activations]{\includegraphics[width=0.248\textwidth,trim=5 5 5 4,clip]{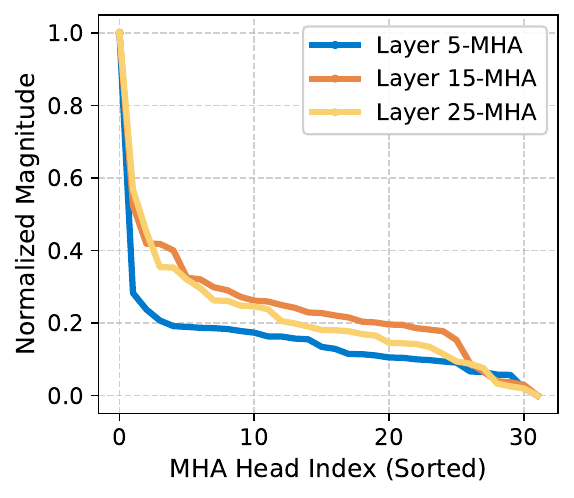}\label{fig:pilot1-mha}}
  \subfloat[FFN Activations]{\includegraphics[width=0.233\textwidth,trim=20 5 5 5,clip]{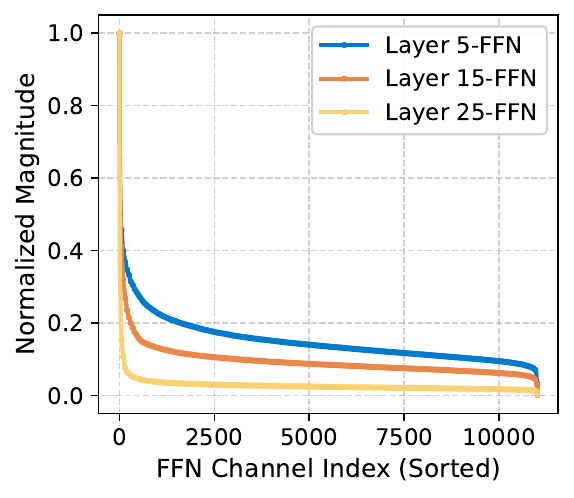}\label{fig:pilot1-ffn}}
  \caption{Normalized activation magnitudes of MHA heads and FFN channels. Both are based on LLaMA2-7B and sorted in descending order.}
  \label{pilot1}
\end{figure}
Motivated by previous findings that only a subset of parameters is crucial for model performance, we are curious about the prevalence of such advantageous parameters across different components, as well as the relationship between the proportion of advantageous parameters and model performance. Following prior work~\cite{sun2024a,DBLP:conf/aaai/AnZYTW24},
we examine the activation magnitudes of MHA heads and FFN channels on a calibration dataset~\cite{ivison2023camelschangingclimateenhancing}. Specifically, we compute the Frobenius norm of the output for each head $h$ and the post-activation output for each channel $c$: $\mathbf{A}^h_{\text{MHA}} = \|\text{head}_h\|_F^2$ and $\mathbf{A}^c_{\text{FFN}} = \|f(\mathbf{X}\mathbf{W}_G^{[:,c]})\|_F^2$.

\begin{figure*}[t]
\centering
\subfloat[7B models-MHA]{ \includegraphics[width=0.258\linewidth,trim=5 5 5 4,clip]{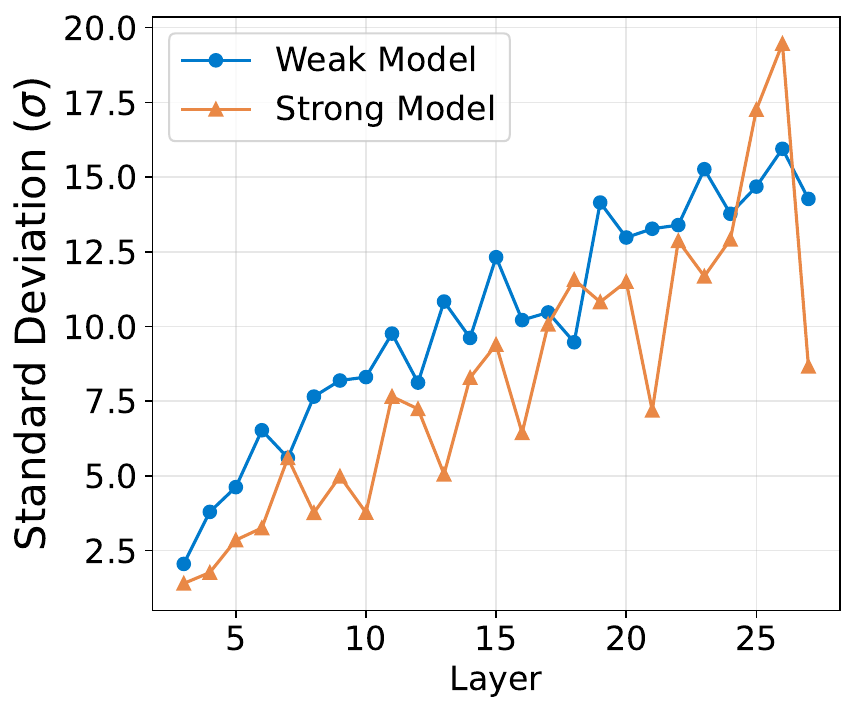}} 
\subfloat[7B models-FFN]{ \includegraphics[width=0.239\linewidth,trim=32 5 5 5,clip]{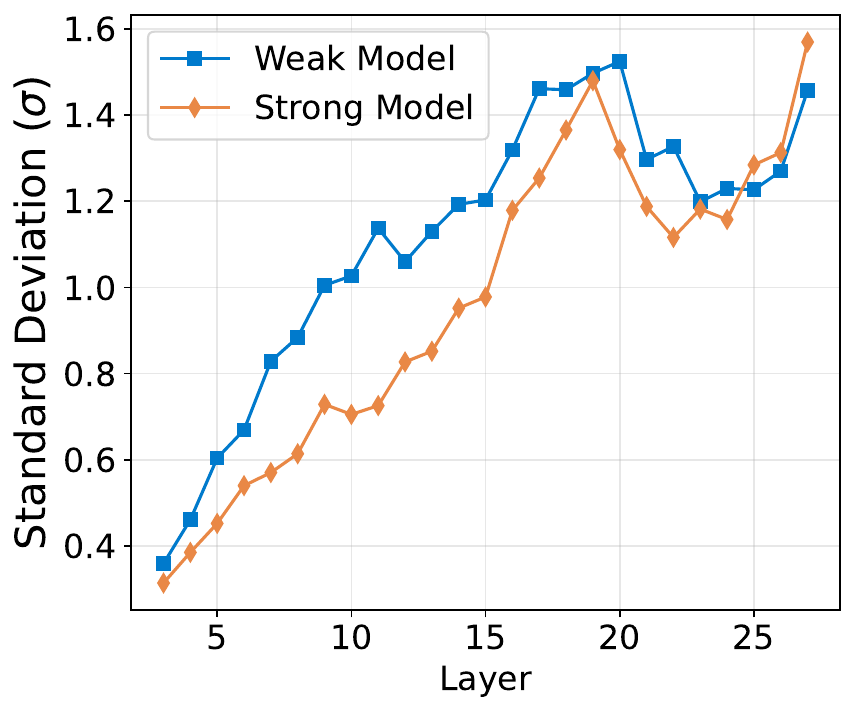}}
\subfloat[3B models-MHA]{	\includegraphics[width=0.239\linewidth,trim=32 5 5 5,clip]{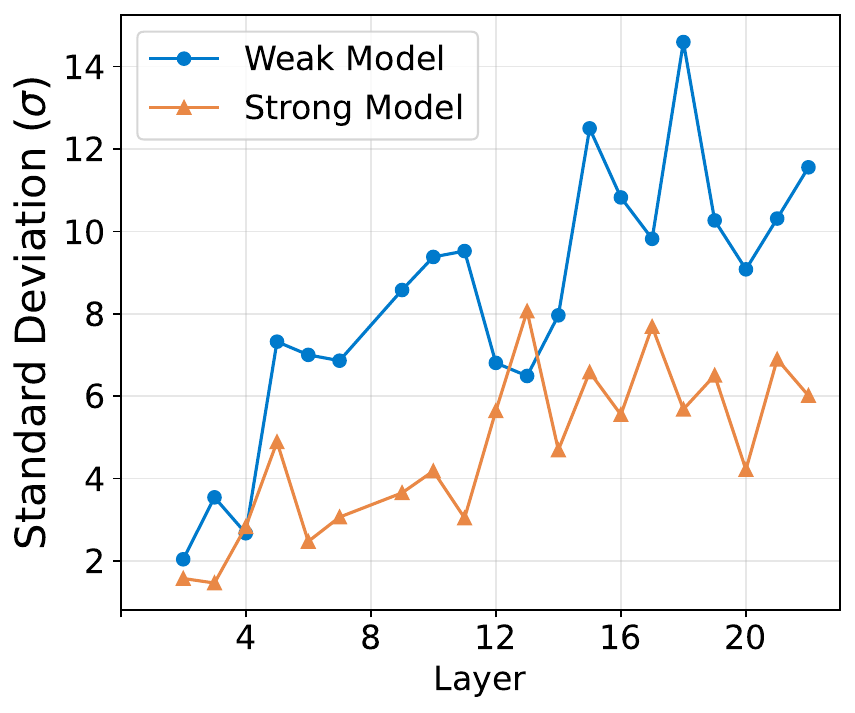}} 
\subfloat[3B models-FFN]{ \includegraphics[width=0.239\linewidth,trim=32 5 5 5,clip]{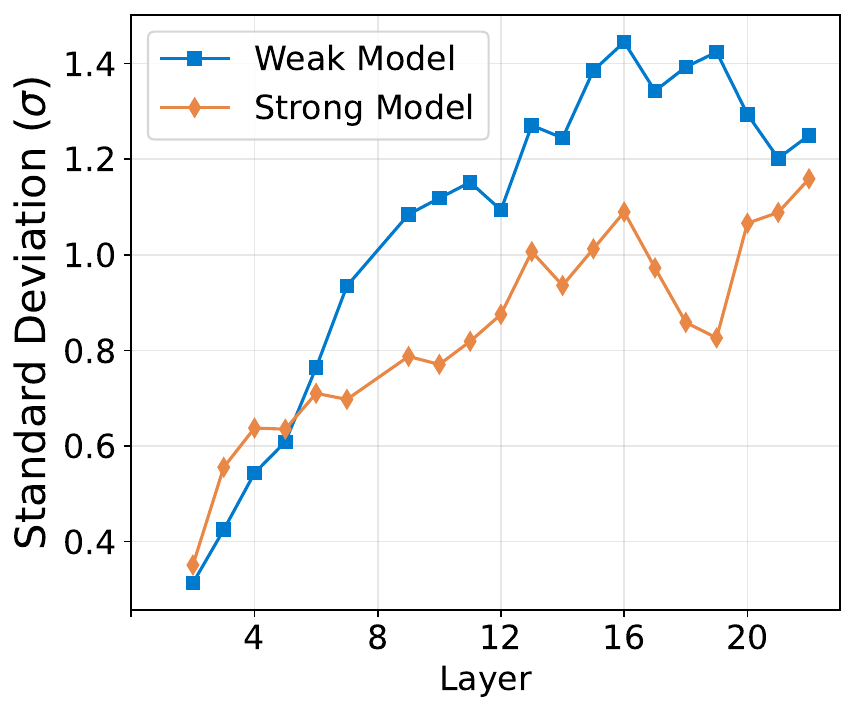}} 
\caption{Activation distribution comparison between models with similar architectures but different performance (i.e., strong vs. weak models). One group consists of standard 7B models trained from scratch~\cite{Touvron2023LLaMAOA,jiang2023mistral7b}. Another group includes 3B models obtained via model compression~\cite{xia2024sheared,grattafiori2024llama3herdmodels}. See Appendix~\ref{app-Imp} for details.}
\label{pilot2}
\end{figure*}
\paragraph{Differences in Parameter Advantages.}
Figures~\ref{fig:pilot1-mha} and~\ref{fig:pilot1-ffn} reveal a distinct long-tailed pattern in activation magnitudes across both MHA and FFN modules. This phenomenon is prevalent across various layers. We further conduct an experiment (Appendix~\ref{app-Exp:obs}) and find that removing parameters with high activations leads to catastrophic degradation in model performance, whereas removing parameters with low activations has little impact. This indicates that across both MHA and FFN modules, a subset of parameters dominates model capability, while the remainder contributes less.

\paragraph{The Relationship Between Advantageous Parameters and Performance.} We further explore the relationship between activation distribution and model capability. We employ standard deviation $\sigma$ to quantify distribution sharpness, where a higher $\sigma$ indicates a more skewed distribution dominated by fewer parameters. As illustrated in Figure~\ref{pilot2}, we compare models of similar architectures but different benchmark performance (i.e., strong vs. weak). We observe that weaker models tend to exhibit sharper activation distributions, suggesting that they possess fewer advantageous parameters while the majority remain underutilized. In contrast, stronger models display more uniform patterns, indicating a broader set of advantageous parameters that contribute more evenly. This motivates our hypothesis that model performance depends not only on parameter count, but on how effectively parameters are utilized.
\section{Method}
\begin{figure*}[t]
    \centering
\includegraphics[width=2\columnwidth,trim=0 60 0 100,clip]{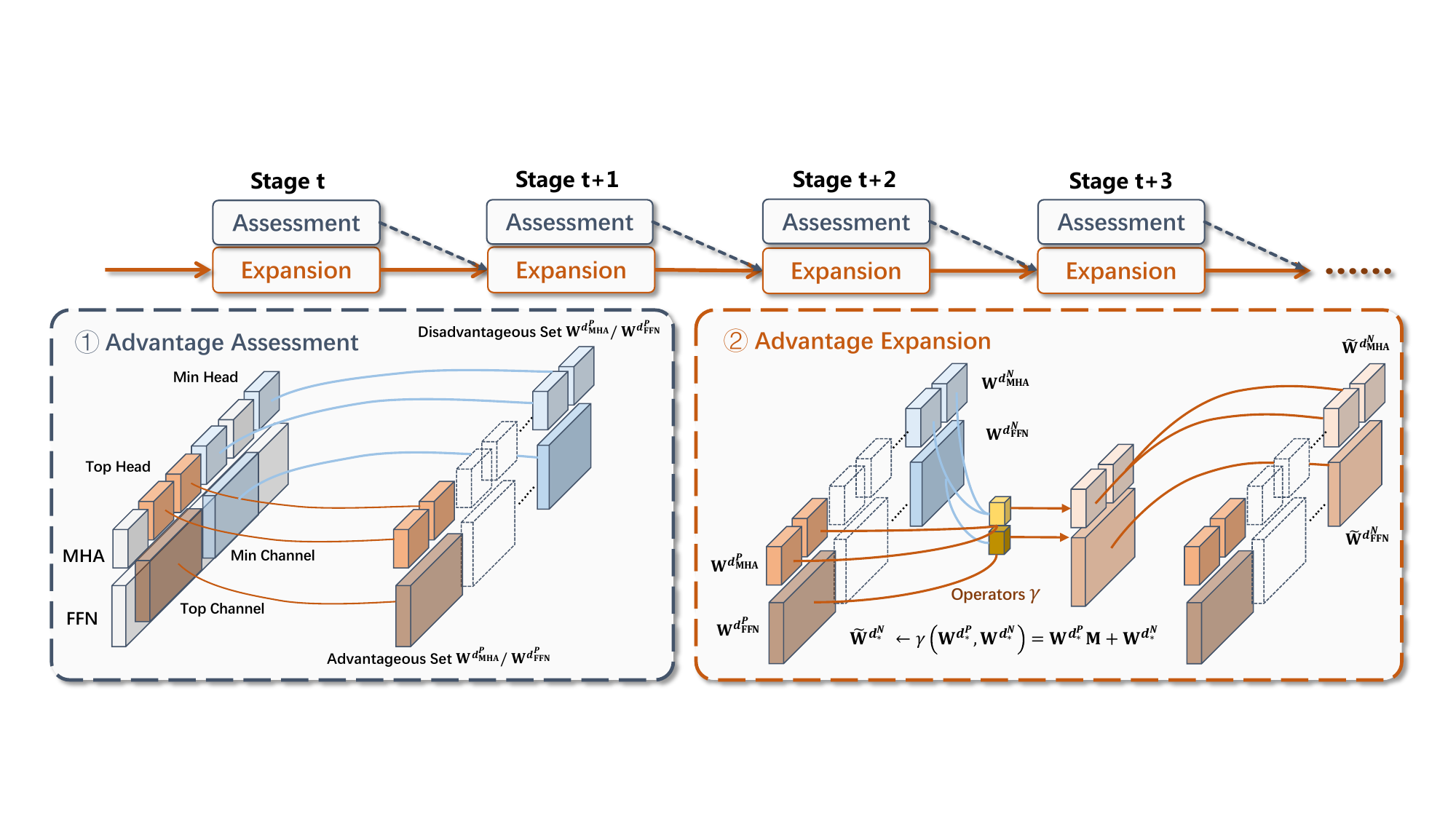}
    \caption{Overview of APEX Training. To progressively expand the 
advantageous parameter space, APEX divides the entire training process into multiple 
stages. At the beginning of each stage, we reselect the sets of advantageous and disadvantageous parameters, and then initialize the new operators based on assessment results collected from the previous stage. In the expansion training of each stage, we continuously perform advantage assessment during each forward pass for the parameter sets selection of the next stage.
    At the end of each stage, we fuse the operators into the model to ensure that the model architecture remains unchanged.}
    \label{fig-method}
\end{figure*}

In this section, we introduce \textbf{A}dvantageous \textbf{P}arameter \textbf{EX}pansion Training (APEX) to increase the proportion of advantageous parameters (Figure~\ref{fig-method}). Sec.~\ref{method-assess} details how we assess advantageous parameters. Sec.~\ref{method-operator} introduces expansion operators for MHA and FFN modules. Sec.~\ref{method-training} describes the stage-wise APEX implementation, and Appendix~\ref{method-theory} provides a theoretical perspective on its mechanism.

\subsection{Advantage Assessment}
\label{method-assess}
As observed in Sec.~\ref{sec-obs}, parameter contributions to model performance are highly uneven. 
To effectively expand the advantages of parameters with higher contributions to those with lower contributions, the primary issue to be addressed is how to accurately assess the advantages of various parameters.
A naive approach is to calculate the average output activations, but it faces two problems, 
(1) the activation magnitudes vary to a relatively large extent across different input samples, and (2) a small number of unusually large activation values may dominate the average, leading to unstable estimates.

Based on the Frobenius norm used in observations, we propose a robust metric with relative ranking. 
For each MHA and FFN module, we calculate the difference between the number of times it appears in the Top-K and Min-K by activation magnitude across the given dataset $\mathcal{D}$. Specifically, for the $h$-th head, the score is computed as:
\begin{equation}
\begin{aligned}
\mathbf{s}_\text{MHA}^h& = \sum_{\mathcal{D}}{\mathbb{I}{[\mathbf{A}^h_\text{MHA} \in \text{Top-K}(\mathbf{A}_\text{MHA})]}}\\&- \sum_{\mathcal{D}}{\mathbb{I}{[\mathbf{A}^h_\text{MHA} \in \text{Min-K}(\mathbf{A}_\text{MHA})]}},
\end{aligned}
\end{equation}
where $\mathbb{I}[\cdot]$ is the indicator function, it returns 1 if the condition inside holds true and 0 otherwise. Unlike simple averaging, this voting-based mechanism filters out the noise from sporadic outliers. Based on the scores, the sets of advantageous parameters and disadvantageous parameters can be selected:
\begin{equation}
d^{P}_{\text{MHA}}=\{h\ |\ \mathbf{s}_\text{MHA}^h \in \text{Top-K}(\mathbf{s}_\text{MHA})\},
\end{equation}
\begin{equation}
     d^{N}_{\text{MHA}}=\{h\ |\ \mathbf{s}_\text{MHA}^h \in \text{Min-K}(\mathbf{s}_\text{MHA})\},
\end{equation}
where $d^{P}_\text{MHA}$ and $d^{N}_\text{MHA}$ denote the dimension indices of the advantageous and disadvantageous parameters set, respectively\footnote{$P$ and $N$ stand for positive (advantageous) and negative (disadvantageous).}. Similarly, we select the sets of advantageous parameters $d^{P}_{\text{FFN}}$ and disadvantageous parameters $d^{N}_{\text{FFN}}$ in the FFN module:
\begin{equation}
\begin{aligned}
\mathbf{s}_\text{FFN}^c& = \sum_{\mathcal{D}}{\mathbb{I}{[\mathbf{A}^c_\text{FFN} \in \text{Top-K}(\mathbf{A}_\text{FFN})]}}\\&- \sum_{\mathcal{D}}{\mathbb{I}{[\mathbf{A}^c_\text{FFN} \in \text{Min-K}(\mathbf{A}_\text{FFN})]}},
\end{aligned}
\end{equation}
\begin{equation}
d^{P}_{\text{FFN}}=\{c\ |\ \mathbf{s}_\text{FFN}^c \in \text{Top-K}(\mathbf{s}_\text{FFN})\}, 
\end{equation}
\begin{equation}
d^{N}_{\text{FFN}}=\{c\ |\ \mathbf{s}_\text{FFN}^c \in \text{Min-K}(\mathbf{s}_\text{FFN})\}.
\end{equation}
For simplicity, we use a unified threshold $K$ for both MHA and FFN throughout all experiments.
\subsection{Advantage Expansion Operators}
\label{method-operator}

\noindent\textbf{MHA Expansion Operators.} Given an advantageous set $\text{head}_P$ and a disadvantageous set $\text{head}_N$, we introduce an expansion operator $\gamma_{\text{MHA}}$ to expand the advantages of $\text{head}_P$ to $\text{head}_N$:
\begin{equation}
\widetilde{\text{head}}_N \leftarrow \gamma_{\text{MHA}}(\text{head}_P,\text{head}_N),
\end{equation}
where the operator $\gamma_{\text{MHA}}$ is realized by applying a learnable transformation to parameter sets $\text{head}_P$ and $\text{head}_N$, resulting in a new set of parameters $\widetilde{\text{head}}_N$ that substitutes for the output of the original parameter set $\text{head}_N$.
Afterwards, considering $\mathbf{W}_O$ receives the concatenated outputs from all heads, its dimensions correspond to those of each head. Therefore, when expanding the heads, the corresponding input dimensions in $\mathbf{W}_O$ should also be expanded to align with $\widetilde{\text{head}}_N$ accordingly:
\begin{equation}
\mathbf{\tilde{W}}_O^{\text{[}d^{N}_{\text{MHA}}\text{,:]}} \leftarrow \gamma_{\text{MHA}}(\mathbf{W}_O^{\text{[}d^{P}_{\text{MHA}}\text{,:]}},\mathbf{W}_O^{\text{[}d^{N}_{\text{MHA}}\text{,:]}}),
\end{equation}

\noindent\textbf{FFN Expansion Operators.} 
Similar to head expansion in the MHA module, we expand advantageous channels onto disadvantageous ones in the FFN module. 
Given the set of dimension indices $d^{P}_{\text{FFN}}$ and $d^{N}_{\text{FFN}}$, we introduce the expansion operator $\gamma_{\text{FFN}}$ to perform this expansion:
\begin{equation}
\mathbf{\tilde{W}}_U^{\text{[:,}d^{N}_{\text{FFN}}\text{]}} \leftarrow \gamma_{\text{FFN}}(\mathbf{W}_U^{\text{[:,}d^{P}_{\text{FFN}}\text{]}},\mathbf{W}_U^{\text{[:,}d^{N}_{\text{FFN}}\text{]}}),
\end{equation}
\begin{equation}
\mathbf{\tilde{W}}_G^{\text{[:,}d^{N}_{\text{FFN}}\text{]}} \leftarrow \gamma_{\text{FFN}}(\mathbf{W}_G^{\text{[:,}d^{P}_{\text{FFN}}\text{]}},\mathbf{W}_G^{\text{[:,}d^{N}_{\text{FFN}}\text{]}}),
\end{equation}
Next, the corresponding input dimensions in $\mathbf{W}_D$ also need to be operated for alignment:
\begin{equation}
\mathbf{\tilde{W}}_D^{\text{[}d^{N}_{\text{FFN}}\text{,:]}} \leftarrow \gamma_{\text{FFN}}(\mathbf{W}_D^{\text{[}d^{P}_{\text{FFN}}\text{,:]}},\mathbf{W}_D^{\text{[}d^{N}_{\text{FFN}}\text{,:]}}).
\end{equation}

\noindent\textbf{Implementation of Operators.} 
Recall that the goal of expansion is to improve the utilization of the
under-utilized (disadvantageous) subspace by propagating directions spanned by advantageous
parameters to it, thereby increasing the number of parameters that behave advantageously during training. To this end, we introduce a transformation matrix $\mathbf{M} \in \mathbb{R}^{d^{P}_*\times d^{N}_*}$ for the matrix $\mathbf{W}^{\text{[:,}d^{P}_*\text{]}}$ to be expanded into the space of $\mathbf{W}^{\text{[:,}d^{N}_*\text{]}}$, where $*$ denotes either MHA or FFN, and $d^{P}_*$, $d^{N}_*$ are the corresponding dimension indices:
\begin{equation}
\begin{aligned}
\tilde{\mathbf{W}}^{\text{[:,}d^{N}_*\text{]}}\leftarrow\gamma&(\mathbf{W}^{\text{[:,}d^{P}_*\text{]}},\mathbf{W}^{\text{[:,}d^{N}_*\text{]}}) \\
&= \mathbf{W}^{\text{[:,}d^{P}_*\text{]}}\mathbf{M}+\mathbf{W}^{\text{[:,}d^{N}_*\text{]}},
\end{aligned}
\end{equation}
where the transformation matrix $\mathbf{M}$ is initialized with zeros to ensure that the starting point of the expansion remains consistent with the original model. As analyzed in Appendix~\ref{method-theory}, the expansion creates an auxiliary optimization path: updates to $\mathbf{M}$ induce an
additional structured term in the effective update of $\tilde{\mathbf{W}}^{\text{[:,}d^{N}_*\text{]}}$ that
is aligned with the span of $\mathbf{W}^{\text{[:,}d^{P}_*\text{]}}$ (see Lemma~\ref{lemma1}). Moreover, since projection does not increase the expected energy of the gradient
noise (Lemma~\ref{lemma2}), this correction can improve the effective signal-to-noise ratio of updates in
under-utilized directions, enabling them to participate more during training.

After the expansion is completed, in order to preserve the original architecture, we fuse the expansion operators into the model: ${\mathbf{W}}^{\text{[:,}d^{N}_*\text{]}} \leftarrow \tilde{\mathbf{W}}^{\text{[:,}d^{N}_*\text{]}}$. Such designs allow APEX to be incorporated into standard training procedures seamlessly. In addition, to decrease the computational complexity of the transformation matrix, we leverage Monarch matrices \cite{dao2022monarchexpressivestructuredmatrices,chen-etal-2024-lemon} to reconstruct it:
\begin{equation}
\begin{aligned}
\mathbf{M}=\left[\begin{array}{ccc}
\mathbf{D}_{1,1}  & \cdots & \mathbf{D}_{1, d} \\
\vdots & \ddots & \vdots \\
\mathbf{D}_{d, 1} & \cdots & \mathbf{D}_{d, d}
\end{array}\right]
&\left[\begin{array}{ccc}
\mathbf{R}_1 & & \\
& \ddots & \\
& & \mathbf{R}_{d}
\end{array}\right],
\end{aligned}
\end{equation}
where $d=\sqrt{\text{min }(d^{P}_*,d^{N}_*)}$ and each $\mathbf{D}$,$\mathbf{R}$ is a $d\times d$ diagonal matrix, thus the mapping complexity is reduced from $O(d^4)$ to $O(d^2)$. After reconstructing the matrix, in order to perform zero initialization, we initialize each $\mathbf{R}$ matrix to zeros and randomly initialize each $\mathbf{D}$ matrix. Appendix~\ref{app-Exp:efficiency} shows the parameter efficiency benefits of the operator.

\subsection{Stage-wise Training}
\label{method-training}
To progressively expand the advantageous parameter space, we divide the entire training process into multiple stages. At the beginning of each stage, we reselect the sets of advantageous and disadvantageous parameters and initialize the new operators based on statistics collected from the previous stage. During the training of each stage, we continuously record metric scores $\mathbf{s}_\text{MHA}$ and $\mathbf{s}_\text{FFN}$ during each forward pass for the parameter sets selection of the next stage\footnote{For the first stage, we identify advantageous parameters by performing forward passes on a small subset of the training set prior to training. In each stage, $\mathcal{D}$ actually refers to the training data of the previous stage.}. At the end of each stage, we fuse the operators into the model:
\begin{equation}
\mathbf{W}_{t}^{\text{[:,}d^{N}_*\text{]}_\text{final}} = \mathbf{W}_{t}^{\text{[:,}d^{P}_*\text{]}}\mathbf{M}_{t}+\mathbf{W}_{t}^{\text{[:,}d^{N}_*\text{]}},\text{ where }t \in T,
\end{equation}
where $T$ denotes the total number of stages. $\mathbf{W}_{t}^{\text{[:,}d^{N}_*\text{]}_\text{final}}$ denotes the parameter values corresponding to the disadvantageous parameter sets after the operators have been fused at stage $t$. Regarding training efficiency, since the assessment is fused into the forward pass and the Monarch-based expansion operator is lightweight, APEX maintains training overhead comparable to other sparse fine-tuning methods (see Appendix~\ref{app-Exp:efficiency} for details).


\section{Empirical Evaluation}
\begin{table*}[t]
\centering
\small
\setlength{\tabcolsep}{3.5pt}
        \begin{tabular}{cccccccc>{\columncolor{gray!20}}cc@{}}
            \toprule
\multicolumn{1}{c}{\multirow{2}{*}{\textbf{Method}}} & \multicolumn{1}{c}{\multirow{2}{*}{\textbf{\#Param.}}} & \textbf{MMLU} & \textbf{GSM8K} & \textbf{BBH} & \textbf{TyDiQA} & \textbf{TruthfulQA} & \textbf{HumanEval} & \\
 &  & \scriptsize0-shot, EM & \scriptsize8-shot CoT, EM & \scriptsize3-shot CoT, EM & \scriptsize1-shot, F1 & \scriptsize0-shot, MC2 & \scriptsize0-shot, Pass@10 &  \multirow{-2}{*}{\textbf{Average}} \\
            \midrule
            Pre-trained$^{\dagger}$ & - & 41.6 & 12.0 & 39.9 & 48.4 & 38.5 & 26.2 & 34.4 \\
            Full-FT & 100\% & 49.6 & 30.8 & 45.2 & 49.4 & 46.9 & 36.5 & 43.1 \\
            \midrule
            \multicolumn{8}{l}{\textit{Parameter-Efficient Fine-tuning}} \\
            LoRA & 2.4\% & 44.1 & 26.5 & 44.1 & 54.0 & 43.7 & 28.8 & 40.2 \\
            DoRA & 2.4\% & 47.7 & 25.5 & 44.0 & 55.0 & 45.4 & 31.8 & 41.6 \\
            PiSSA & 2.4\% & 48.2 & 25.5 & 42.5 & 50.6 & 45.5 & 33.6 & 41.0 \\
            ReLoRA & 2.4\% & 47.4 & 25.5 & 43.4 & 55.1 & 43.9 & 31.4 & 41.1 \\
            S$^{2}$FT & 2.4\% & 47.6 & 25.0 & 45.5 & 52.8 & 43.7 & 33.9 & 41.4 \\
            APEX & 2.4\% & 48.1 & 28.5 & 45.7 & 54.5 & 45.5 & 33.2 & \textbf{42.6} \\
            \midrule
            \multicolumn{8}{l}{\textit{Half Parameter Fine-tuning}} \\
            HFT$^{\dagger}$ & 52\% & 50.8 & 30.5 & 43.6 & 52.3 & 45.4 & 34.6 & 42.9 \\
            APEX & 52\% & 49.5 & 33.5 & 46.5 & 54.0 & 46.3 & 34.2 & \textbf{44.0} \\
            \cmidrule(lr){1-9}
            RMT$^{\ddagger}$ & 60\% & 47.4 & 34.5 & 44.4 & 52.9 & 47.9 & 33.7 & 43.4 \\
            GMT$^{\ddagger}$ & 60\% & 47.6 & 38.0 & 43.3 & 53.1 & 47.5 & 34.6 & 44.0 \\
            APEX & 60\% & 50.0 & 34.5 & 46.4 & 53.9 & 45.9 & 36.6 & \textbf{44.6} \\
            \bottomrule
        \end{tabular}
\caption{Instruction Tuning results for LLaMA2-7B. \#Param. denotes trainable parameters. The results are averaged over three runs. All results with $^{\dagger}$ are taken from HFT~\cite{hui-etal-2025-hft} and those marked with $^{\ddagger}$ are taken from GMT~\cite{DBLP:conf/aaai/LiZLGWCC25}. Our experiments are conducted under the exact same settings as these works. More results on LLaMA3.1-8B, LLaMA2-13B, and Mistral-7B-v0.1 are provided in Appendix~\ref{app-Exp:models}.}
\label{results-it}
\end{table*}
We conduct experiments under two scenarios: instruction tuning and continued pre-training. For instruction tuning, we compare methods across different trainable parameter settings. For continued pre-training, we assess APEX’s improvements in convergence speed and performance on downstream tasks over conventional training. 

\subsection{Instruction Tuning}
\label{sec-it}
\paragraph{Models and Data.}
To comprehensively evaluate the performance of different methods, following~\citet{hui-etal-2025-hft}, we use \textsc{T\"ulu}~V2 instruction-tuning dataset~\cite{ivison2023camelschangingclimateenhancing} as the training set, which contains a variety of high-quality data. We employ LLaMA2-7B~\cite{touvron2023llama2openfoundation}, LLaMA3.1-8B~\cite{grattafiori2024llama3herdmodels}, LLaMA2-13B, and Mistral-7B-v0.1~\cite{jiang2023mistral7b} as base models. The evaluation covers several aspects: MMLU~\cite{hendrycks2021measuring} (factual knowledge), GSM8K~\cite{DBLP:journals/corr/abs-2110-14168} and BBH~\cite{suzgun-etal-2023-challenging} (reasoning), TyDiQA~\cite{clark-etal-2020-tydi} (multilinguality), TruthfulQA~\cite{lin-etal-2022-truthfulqa} (truthfulness), HumanEval~\cite{DBLP:journals/corr/abs-2107-03374} (coding).

\noindent\textbf{Baselines and Settings.}
We compare APEX with two baseline categories. The first includes parameter-efficient fine-tuning~(PEFT) methods 
with a minimal number of trainable parameters. Specifically, we compare against widely used approaches LoRA~\cite{hu2022lora}, as well as its variants DoRA~\cite{liu2024dora}, ReLoRA~\cite{lialin2024relora}, PiSSA~\cite{meng2024pissa}, as well as the structured sparse fine-tuning method S$^{2}$FT~\cite{yang2024s2ft}. Among these methods, S$^{2}$FT serves as a closely related baseline. It directly fine-tunes selected structured submatrices, whereas APEX learns zero-initialized expansion operators to expand advantageous parameters into the space of disadvantageous ones and fuses these operators into the model after each training stage.
The second category consists of half parameter fine-tuning methods that enable approximately half of the model parameters to be trainable, including HFT~\cite{hui-etal-2025-hft}, which randomly selects half of the weight matrices to be trainable; RMT and GMT~\cite{DBLP:conf/aaai/LiZLGWCC25}, which update parameters selected randomly or according to gradient magnitude, respectively. APEX controls the number of trainable parameters by allowing only the selected parameter set and expansion operators to be trainable.
More details are presented in Appendix~\ref{app-Imp}.

\noindent\textbf{Results.}
Table~\ref{results-it} reports the results of instruction tuning. Compared with other PEFT methods and half-parameter tuning approaches under various trainable parameter settings, APEX consistently achieves superior average performance. Specifically, in the PEFT setting, APEX achieves an average improvement of over 1\% across multiple tasks compared to other methods. Notably, it uses only 2.4\% of trainable parameters, yet approaches the performance of HFT, which requires training half of the model parameters. In the half-parameter tuning methods, APEX  (with 52\% parameters) clearly outperforms HFT and Full-FT, while matching the performance of GMT, which requires 60\%. Furthermore, when the budget is increased to 60\%, APEX pushes the model capability to a new performance peak. These results demonstrate the effectiveness of advantage expansion. More results on other models are provided in Appendix~\ref{app-Exp:models}. Overall, APEX consistently demonstrates superior performance across various models.

\begin{table*}[t]
\setlength\tabcolsep{4pt}
\footnotesize
    \centering
    \begin{tabular}{c ccc cccccc cc ccc@{}}
    \toprule
    \multirow{2}{*}{\raisebox{-0.5\height}{\textbf{Method}}} & \multirow{2}{*}{\raisebox{-0.5\height}{\textbf{\#Param.}}} & \multirow{2}{*}{\raisebox{-0.5\height}{\textbf{PPL($\downarrow$)}}} & \multicolumn{6}{c}{\scriptsize\textbf{Commonsense \& Reading Comprehension}} & \multicolumn{2}{c}{\scriptsize\textbf{Continued}} &  \multicolumn{1}{c}{\scriptsize\textbf{LM}} & \multicolumn{1}{c}{\scriptsize\textbf{Knowledge}} & {\multirow{2}{*}{\raisebox{-0.5\height}{\textbf{Avg.}}}} \\
    \cmidrule(lr){4-9} \cmidrule(lr){10-11} \cmidrule(lr){12-12} \cmidrule(lr){13-13}
    & & & \scriptsize\textbf{SciQ} & \scriptsize\textbf{PIQA} & \scriptsize\textbf{WG} & \scriptsize\textbf{ARC-E } & \scriptsize\textbf{ARC-C} & \scriptsize\textbf{Hella.} & \scriptsize\textbf{LogiQA}  & \scriptsize\textbf{BoolQ} & \scriptsize\textbf{Lam.} & \scriptsize\textbf{MMLU} &  \\
    \midrule
    \multicolumn{6}{l}{\textit{Training budget from $~0B$ to $~3B$}} \\
    Vanilla & 100\% & 5.87 & 87.9 & 71.8 & 57.1 & 58.8 & 28.0 & 44.1 & 20.0 & 63.9 & 58.2 & 24.9 & \cellcolor{gray!20}{51.5} \\
    APEX & 100\% & 5.75 & 88.4 & 71.9 & 57.5 & 60.8 & 29.1 & 44.7 & 21.4 & 63.9 & 58.6 & 25.5 & \cellcolor{gray!20}{\textbf{52.2}} \\
    \midrule
    \multicolumn{6}{l}{\textit{Training budget from $~3B$ to $~6B$}} \\
    Vanilla & 100\% & 5.85 & 88.1 & 72.2 & 58.6 & 58.5 & 28.6 & 44.0 & 20.7 & 62.7 & 57.8 & 25.1 & \cellcolor{gray!20}{51.6} \\
    APEX & 100\% & 5.55 & 88.3 & 72.4 & 58.8 & 61.0 & 29.4 & 45.7 & 21.0 & 64.2 & 60.2 & 26.1 & \cellcolor{gray!20}{\textbf{52.7}} \\
    \midrule
    \multicolumn{6}{l}{\textit{Training budget from $~6B$ to $~10B$}} \\
    Vanilla & 100\% & 5.83 & 86.6 & 71.9 & 58.6 & 56.8 & 33.2 & 44.4 & 23.2 & 62.1 & 58.8 & 25.3 & \cellcolor{gray!20}{52.1} \\
    APEX & 100\% & 5.44 & 88.5 & 73.1 & 59.6 & 61.6 & 29.9 & 45.9 & 20.9 & 62.9 & 61.4 & 26.9 & \cellcolor{gray!20}{\textbf{53.1}} \\
    \bottomrule
    \end{tabular}
\caption{Continued pre-training results for APEX and conventional training (Vanilla). The results are averaged over three runs. In APEX, we reselect the advantage and disadvantage parameter sets every 3B data. And the initial point of the next stage is based on the checkpoint of the previous stage.}
\label{table:pt}
\end{table*}

\subsection{Continued Pre-Training}
\label{sec-pt}

\noindent\textbf{Models and Data.} We utilize TinyLLaMA-v1-1.1B~\cite{zhang2024tinyllamaopensourcesmalllanguage} as the base model and use the RedPajama dataset~\cite{weber2024redpajama} for continued pre-training. The training corpus consists of 10 billion tokens, accompanied by a validation set with 2 million tokens. For evaluation, we follow the procedures established by LLaMA2 and report accuracy across 10 representative downstream tasks: ARC-Easy/Challenge~\cite{clark2018think}, LAMBADA~\cite{paperno2016lambada}, LogiQA~\cite{liu2020logiqa}, PIQA~\cite{DBLP:conf/aaai/BiskZLGC20}, SciQ~\cite{welbl-etal-2017-crowdsourcing}, WinoGrande~\cite{WinoGrande:conf/aaai/SakaguchiBBC20}, HellaSwag~\cite{HellaSwag:conf/acl/ZellersHBFC19}, MMLU, and BoolQ~\cite{clark2019boolq}.

\noindent\textbf{Baselines and Settings.} We adopt the conventional continued pre-training of TinyLLaMA (Vanilla) as our primary baseline. For APEX, we first assess parameter advantages using 30 million tokens, then perform continued pre-training on 10 billion tokens. The baseline is trained on the same 10 billion tokens for fair comparison.

\noindent\textbf{Results.} 
Table~\ref{table:pt} presents the performance after continued pre-training with APEX across multiple stages. The results show that APEX achieves a significant reduction in language modeling perplexity (PPL).
Specifically, compared to Vanilla CPT, which attains a perplexity of 5.83 after training on 10B tokens, APEX reduces the perplexity to 5.75 after training on only 3B tokens, achieving a lower perplexity with only approximately 30\% of the training data. After training on 10B tokens, APEX further reduces the perplexity to 5.44, representing a decrease of approximately 0.39. Notably, the improvement becomes more pronounced as training progresses: the PPL gap between APEX and Vanilla widens from 0.12 at 3B, to 0.30 at 6B, and to 0.39 at 10B, suggesting that advantage expansion provides cumulative benefits. Moreover, the APEX model consistently outperforms the baseline on most of the downstream tasks. These results indicate that advantage expansion effectively enhances the model’s knowledge storage and reasoning capabilities.

\begin{table}[t]
\centering
\small
\setlength{\tabcolsep}{3pt}
\scalebox{0.95}{
\begin{tabular}{@{}lcccccc>{\columncolor{gray!20}}cc@{}}
\toprule
\textbf{Method} & \textbf{MML.} & \textbf{GSM.} & \textbf{BBH} & \textbf{TyD.} & \textbf{Tru.} & \textbf{Hum.} & \textbf{Avg.}  \\
\midrule
HFT-52\% & 50.8 & 30.5 & 43.6 & 52.3 & 45.4 & 34.6 & 42.9 \\
   \midrule
 APEX-52\% & 49.5 & 33.5 & 46.5 & 54.0 & 46.3 & 34.2 & 44.0 \\
\ \ \ \textit{w/o Exp.} & 48.7 & 31.0 & 44.4 & 54.6 & 46.0 & 32.7 & 42.9 \\
\ \ \ \textit{w/o Ass.} & 49.9 & 30.5 & 46.3 & 53.2 & 45.3 & 34.2 & 43.2 \\
\ \ \ \textit{w/o Mon.} & 49.5 & 32.0 & 46.8 & 53.1 & 45.3 & 36.0 & 43.8 \\
 \bottomrule
\end{tabular}
}
\caption{Ablation experiments of APEX.
\textit{w/o Exp.} denotes removing expansion and only finetuning the advantageous parameters;
\textit{w/o Ass.} denotes removing assessment and performing random expansion;
\textit{w/o Mon.} denotes replacing the Monarch layer with a dense layer.}
\label{table-ablation}
\end{table}

\subsection{Analysis}
\label{sec-analysis}

\noindent \textbf{Ablation Study.} Table~\ref{table-ablation} presents the ablation study of our method. Removing either Advantage Expansion or Advantage Assessment significantly degrades performance, confirming that both components are essential.
Furthermore, replacing the Monarch matrix with a fully dense linear operator also reduces performance. We attribute this to the dense matrix's 
excessive degrees of freedom, which introduce noise and instability 
during fine-tuning with limited steps. In contrast, the structured 
Monarch form provides a more stable reparameterization. See Appendix~\ref{app-Exp:analysis} for extended analysis.

\begin{figure}[t]
    \subfloat[Effect of stages $T$.]{ \includegraphics[width=0.50\linewidth,trim=5 20 5 5,clip]{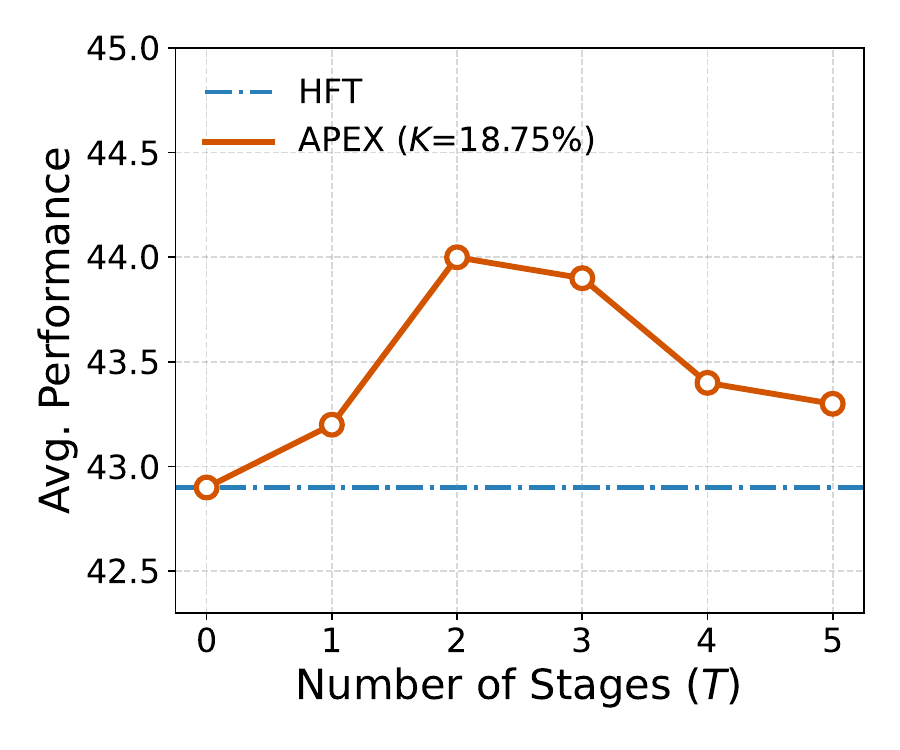}\label{fig:impact-stage}}
    \subfloat[Effect of threshold $K$.]{	\includegraphics[width=0.47\linewidth,trim=5 5 5 5,clip]{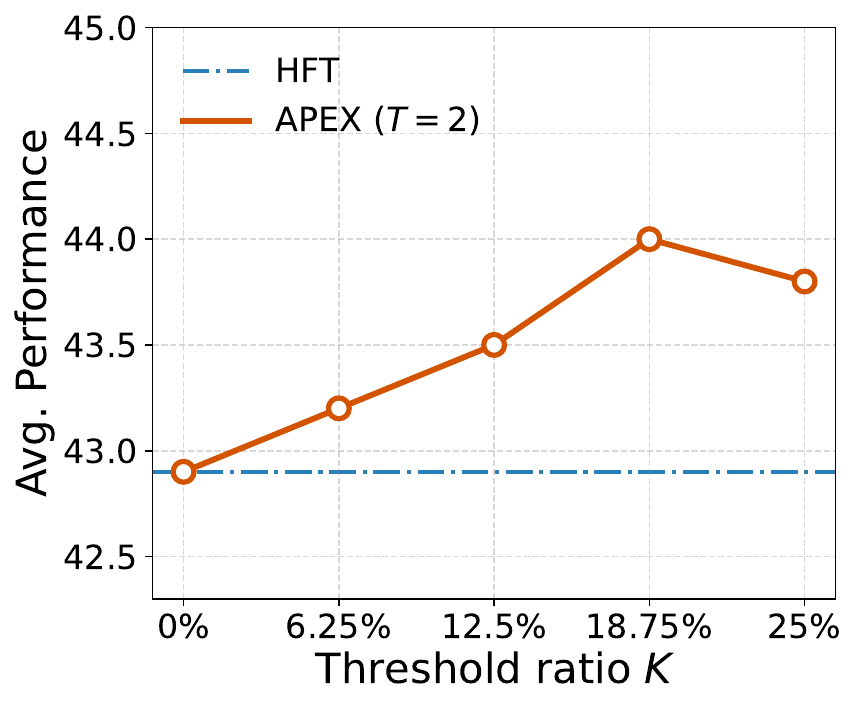}\label{fig:impact-k}}
\caption{Effect of the number of stages $T$ and the threshold ratio $K$. The threshold $K$ is swept up to 25\% under the 50\% half-parameter tuning budget.}
\end{figure}
\begin{figure*}[t]
\centering
\subfloat[MHA (SFT)]{	\includegraphics[width=0.258\linewidth,trim=5 5 5 4,clip]{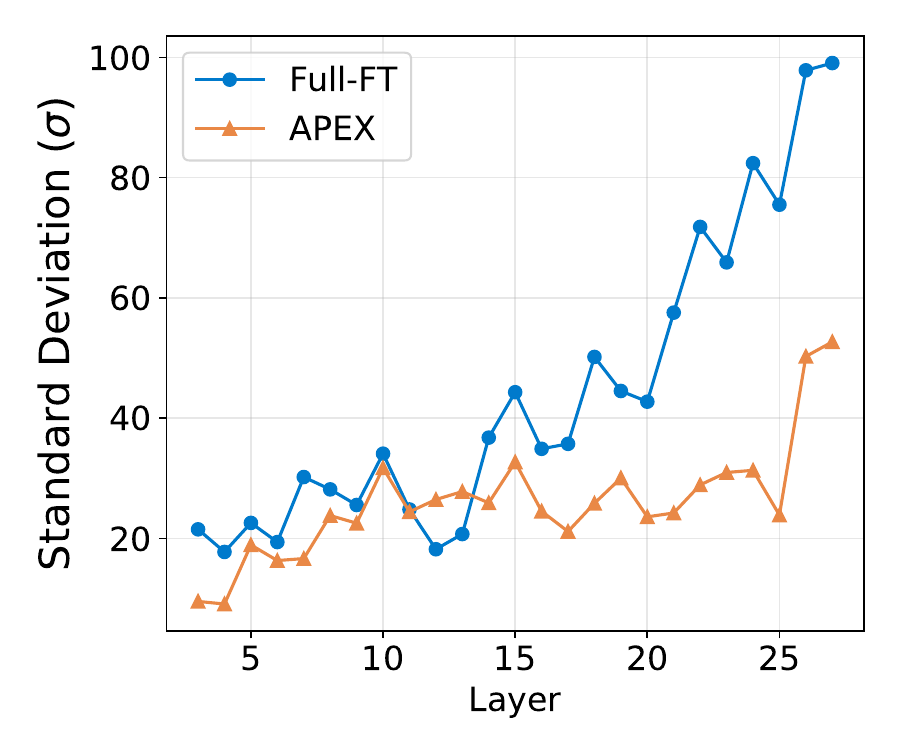}\label{fig-attn-act-dist-sft}}
\subfloat[FFN (SFT)]{ \includegraphics[width=0.239\linewidth,trim=43 5 5 5,clip]{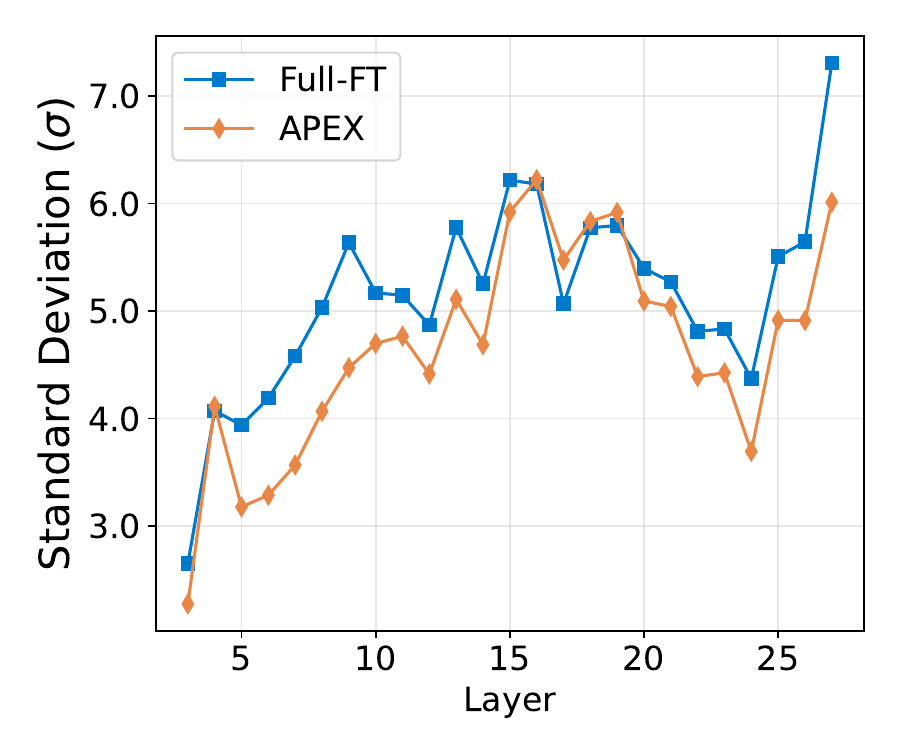}\label{fig:std-sft}}
\subfloat[MHA (CPT)]{ \includegraphics[width=0.239\linewidth,trim=43 5 5 5,clip]{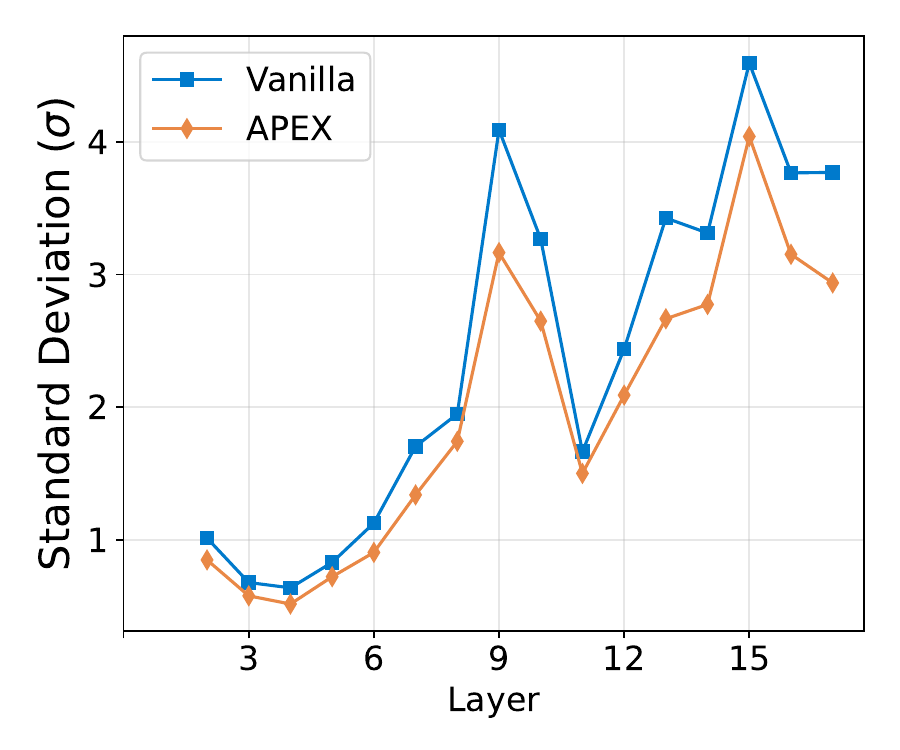}\label{fig-attn-act-dist-cpt}}
\subfloat[FFN (CPT)]{ \includegraphics[width=0.239\linewidth,trim=43 5 5 5,clip]{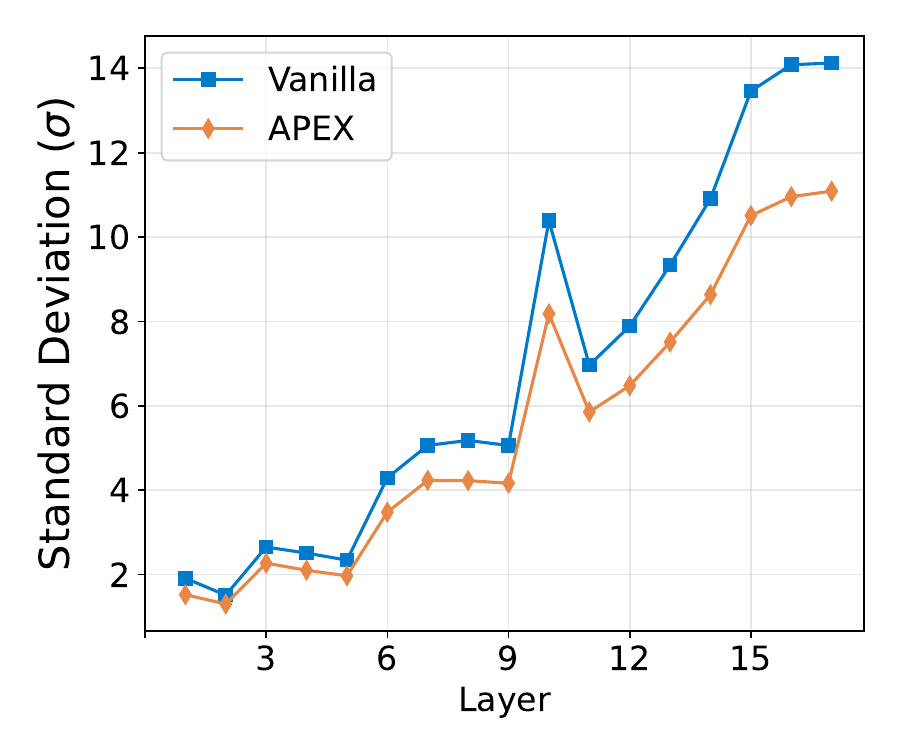}\label{fig:std-cpt}}
\caption{Comparison of activation distributions between models trained with APEX and those trained with conventional methods under both instruction tuning (SFT) and continued pre-training (CPT) scenarios.}
\label{fig:apex_vs_vanilla}
\end{figure*}

\noindent \textbf{Effect of the number of stages $T$.}
Figure~\ref{fig:impact-stage} shows that performance improves as $T$ increases from 0 to~2, indicating that expansion–merge cycles effectively enhance learning. The performance peaks at $T=2$–$3$, which is close to the number of training epochs, while larger $T$ yields no further benefit. We attribute this to the fact that, under a fixed total step budget, excessive stages result in overly brief training intervals. This leads to insufficient data exposure per stage, which hampers the accurate estimation of advantageous directions for parameter expansion. Consequently, we set $T$ equal to the number of training epochs, which naturally aligns the expansion frequency with the data exposure in our experiments. Corresponding analysis for continued pre-training is provided in Appendix~\ref{app-Exp:analysis}.

\noindent \textbf{Effect of the threshold $K$.} Figure~\ref{fig:impact-k} shows the impact of increasing the selection threshold $K$ on half-parameter tuning performance, where we sweep $K$ up to 25\% since half-parameter tuning caps the trainable budget at 50\% (i.e., $2K\le 50\%$). When $K=0$, no expansion occurs, and APEX performs similarly to HFT. As $K$ increases, performance improves steadily, peaking at $K=18.75\%$. Further increasing $K$ yields diminishing returns, likely because expanding too many dimensions enlarges the transformation subspace and makes optimization harder without introducing additional useful advantageous directions. Accordingly, for settings with a trainable budget below 50\% (e.g., PEFT), we set $K$ to half of the trainable ratio, while for budgets above 50\%, we fix $K=18.75\%$.

\noindent \textbf{Observations.} Figure~\ref{fig:apex_vs_vanilla} visualizes the standard deviations of MHA and FFN activations in both instruction tuning and continued pre-training scenarios. Models trained with APEX consistently exhibit lower standard deviations compared to conventional training, indicating more balanced activation distributions and broader parameter utilization. This aligns with our observation in Section~\ref{sec-obs} that stronger models exhibit flatter distributions, confirming that APEX effectively increases the proportion of advantageous parameters.
\section{Related Work}
\noindent \textbf{Parameter-Efficient Training.}
\label{related-work-sparse-ft}
The goal of Parameter-Efficient Fine-Tuning (PEFT) is to adapt large models with minimal trainable parameters. Existing approaches effective for LLMs generally fall into two categories: addition-based and partial parameter fine-tuning methods. Among addition-based methods, LoRA~\cite{hu2022lora} stands out for its efficiency. Recent variants like DoRA~\cite{liu2024dora} and PiSSA~\cite{meng2024pissa} further optimize the mechanism. Partial parameter methods identify and update only a subset of the original parameters. While early works~\cite{guo-etal-2021-parameter,sung2021training} focused on small models, recent approaches like S$^{2}$FT~\cite{yang2024s2ft} have scaled to LLMs. Despite these advances, some studies~\cite{ivison2023camelschangingclimateenhancing,dou-etal-2024-loramoe,gu-etal-2025-beamlora} show that PEFT techniques may underperform full-parameter fine-tuning on complex tasks. To bridge this gap, high-ratio tuning methods have emerged: HFT~\cite{hui-etal-2025-hft} randomly fine-tunes half of the parameters, while GMT~\cite{DBLP:conf/aaai/LiZLGWCC25} employs gradient-based selection. Different from existing approaches, APEX introduces a novel paradigm. By training only operators for expansion and the expanded advantageous and disadvantageous parameters, APEX achieves superior efficacy across various parameter budgets.

\noindent \textbf{Activations in Large Language Models.}
Activation magnitude reflects the strength of the signal propagated through a model and thus provides a natural, data-dependent proxy for how strongly the associated parameters contribute to model outputs~\cite{muralidharan2024compactlanguagemodelspruning,lin2024awqactivationawareweightquantization}. Accordingly, activation-based importance estimation has been widely used in model compression. For instance, Wanda~\cite{sun2024a} and FLAP~\cite{DBLP:conf/aaai/AnZYTW24} prune up to 50\% of parameters based on activation metrics without retraining. Collectively, these studies reveal inherent redundancy in LLMs, implying substantial underutilized capacity. Unlike pruning methods that discard such parameters, APEX assesses parameter advantage across MHA heads and FFN channels and treats underutilized parameters as available resources. By expanding advantageous parameters into the space of disadvantageous ones, APEX improves parameter utilization without altering the final model architecture.

\section{Conclusion}
In this paper, we introduce \textbf{A}dvantageous \textbf{P}arameter \textbf{EX}pansion Training (APEX), a novel paradigm designed to maximize parameter utilization without scaling up model size. Specifically, APEX identifies advantageous parameters and progressively expands them into the space of disadvantageous parameters, thereby increasing their proportion and improving model capabilities. Extensive experiments in both instruction tuning and continued pre-training scenarios show that APEX can achieve the performance of models with more trainable parameters, thus reducing computational costs. Crucially, APEX preserves the original architecture of the model and can be seamlessly integrated into standard training procedures, offering research value and broad application potential.
\section*{Limitations}
APEX demonstrates its effectiveness in training scenarios that build upon pre-existing model parameters, such as continued pre-training or fine-tuning processes. However, for scenarios where pre-training is conducted entirely from scratch, it is necessary to first conduct conventional training for a period before transitioning to APEX training. Our future work will investigate how to better extend APEX to support this scenario.
\section*{Ethical Considerations}
In this study, we employ publicly available datasets and techniques to avoid privacy concerns. These resources are used solely to investigate APEX, our proposed efficiency-focused algorithm, with the aim of reducing training costs for a wide range of model users, thus offering a positive social impact. To minimize the potential impact of data bias on our experimental results, we select large-scale and diverse datasets for our experiments. In addition, we strictly adhered to relevant laws, regulations, and ethical guidelines throughout the research process. All results and methodologies are reported transparently to facilitate reproducibility and further ethical evaluation by the research community.

\section*{Acknowledgments}
This work was supported by the National Natural Science Foundation of China (Nos. 62472419, 62472420) and the Beijing Nova Program
(Grant No. 20250484895).

\bibliography{custom}

\appendix
\newpage
\appendix
\begin{center}
    \Large \textbf{Appendix}
\end{center}
\section{Implementation Details}
\label{app-Imp}
\subsection{Observations}
In our observational experiments, we investigate the relationship between the proportion of advantageous parameters and model performance by examining two types of base models when selecting strong and weak candidates. The first type comprises standard pre-trained models; we use LLaMA-7B~\cite{Touvron2023LLaMAOA} and Mistral-7B~\cite{jiang2023mistral7b} for comparison, as they possess highly similar architectural designs, with empirical results demonstrating that the latter significantly outperforms the former~\cite{open-llm-leaderboard-v2}. The second type consists of efficient small models obtained by compressing standard base models. Specifically, we select Sheared-LLaMA-2.7B~\cite{xia2024sheared}, pruned from LLaMA2~\cite{touvron2023llama2openfoundation}, and LLaMA3.2-3B, obtained from the larger LLaMA3 model. Our activation statistics are based on inference performed on 2048 samples with the same fixed random seed in GSM8K~\cite{DBLP:journals/corr/abs-2110-14168}. For each component, we average the activations across samples and compute the standard deviation of these averaged activations across different heads in each MHA module and across different channels in each FFN module.
\subsection{Instruction Tuning}
In our instruction tuning experiments, we use LLaMA2-7B, LLaMA2-13B~\cite{touvron2023llama2openfoundation}, LLaMA3.1-8B~\cite{grattafiori2024llama3herdmodels}, and Mistral-7B-v0.1~\cite{jiang2023mistral7b} as our backbone models. We consider strong baselines under different trainable parameter settings for instruction fine-tuning, including parameter-efficient fine-tuning, half-parameter fine-tuning, and full-parameter fine-tuning for comparison.

\paragraph{Baselines.} 
For PEFT methods, we select LoRA~\cite{hu2022lora}, which simulates full-rank model updates using a low-rank matrix. We also include three subsequent improvements to LoRA: DoRA~\cite{liu2024dora}, which decomposes LoRA modules into magnitude and direction components to effectively enhance LoRA’s learning capacity, and ReLoRA~\cite{lialin2024relora}, which repeatedly merges trained LoRA parameters back into the base model and reinitializes the LoRA modules, thereby increasing the model’s capacity, and PiSSA~\cite{meng2024pissa}, which utilizes principal singular values to initialize the LoRA modules. All of their ranks are set to 64. Additionally, we select the S$^{2}$FT~\cite{yang2024s2ft} method, which randomly trains either a row or a column of weights in the base model. For both S$^{2}$FT and our proposed method, APEX, we set the number of trainable parameters to match that of rank=64 in LoRA-based methods. For half-parameter fine-tuning methods, HFT~\cite{hui-etal-2025-hft} randomly selects half of the weight matrices along with all embedding and LM head layers for updating, resulting in 52\% trainable parameters. RMT~\cite{DBLP:conf/aaai/LiZLGWCC25} randomly masks 40\% of the gradients during full-parameter fine-tuning, while GMT~\cite{DBLP:conf/aaai/LiZLGWCC25} selects gradients based on their importance, both resulting in 60\% trainable parameters. APEX follows the HFT setting, keeping the trainable parameter ratio at 52\%.

\paragraph{Training Data.} 
We use the \textsc{T\"ulu}~V2 dataset~\cite{ivison2023camelschangingclimateenhancing} for training. This dataset consists of a wide range of sources to enhance the model’s instruction-following capabilities. Specifically, it includes data from FLAN v2~\cite{chung2022scalinginstructionfinetunedlanguagemodels}, Open Assistant~\cite{köpf2023openassistantconversationsdemocratizing}, ShareGPT\footnote{https://sharegpt.com/}, GPT4-Alpaca~\cite{peng2023instructiontuninggpt4}, Code-Alpaca~\cite{codealpaca}, LIMA~\cite{zhou2023limaalignment}, WizardLM Evol Instruct~\cite{xu2023wizardlmempoweringlargelanguage}, Open-Orca~\cite{OpenOrca}, Hardcoded, and Science.

\paragraph{Experimental Details.}
Under the half-parameter fine-tuning setting, we follow Hui et al.~\cite{hui-etal-2025-hft} and set the learning rate to 2e-5 for LLaMA2-7B/13B and 5e-6 for LLaMA3.1-8B and Mistral-7B-v0.1, consistently across all methods. Under the parameter-efficient fine-tuning setting, we apply LoRA to all linear layers for LoRA-based baselines. We select the best learning rate per method from \{1e-4,2e-4\} for LLaMA2-family models following~\cite{yang2024s2ft}; for LLaMA3.1-8B and Mistral-7B-v0.1, we tune the learning rate from \{2e-5,3e-5\} and \{8e-6,1e-5,2e-5\}, respectively. All models are fine-tuned on the dataset for 2 epochs with a batch size of 128. All experiments are conducted on 8 NVIDIA H100 GPUs (80GB each).
\paragraph{Evaluation.}
We follow the evaluation procedures established by~\citet{hui-etal-2025-hft}. Specifically, we report 0-shot accuracy on MMLU~\cite{hendrycks2021measuring} to assess the model’s understanding of factual knowledge. For GSM8K~\cite{DBLP:journals/corr/abs-2110-14168} and BBH~\cite{suzgun-etal-2023-challenging}, we report 8-shot and 3-shot exact match, respectively, to evaluate reasoning abilities. We use F1 score on TyDiQA~\cite{clark-etal-2020-tydi} to measure multilingual capabilities, and the multi-true score (MC2) on TruthfulQA~\cite{lin-etal-2022-truthfulqa} to evaluate the ability to generate truthful information. For code generation, we report pass@10 performance on HumanEval~\cite{DBLP:journals/corr/abs-2107-03374}.

\subsection{Continued Pre-Training} 
In our pre-training experiments, we use the TinyLLaMA-v1-1.1B model~\cite{zhang2024tinyllamaopensourcesmalllanguage} as the source model. After evaluating the advantages of APEX operators on 30M tokens, we proceed with pre-training on 10B tokens. For direct comparison, we also pre-train the original model on 10B tokens as a baseline. All training is conducted with full-parameter training.

\paragraph{Baseline.} We use continued pre-training of TinyLLaMA v1 1.1B as our primary baseline (Vanilla CPT). The original work provides a robust training framework, dynamic batch gradient pre-training strategies, and thorough evaluation protocols, resulting in convincing and competitive performance.
\paragraph{Training Data.} We use the RedPajama dataset~\cite{weber2024redpajama} for both APEX training and conventional training. This dataset mirrors the LLaMA1 pre-training data, covering seven domains: CommonCrawl, C4, Github, Wikipedia, Books, ArXiv, and Stack-Exchange. Consistent with LLaMA’s settings, we use a sequence length of 4096. Our data splits include a 2M tokens validation set, a 4B tokens training set, and a 10B tokens continual pre-training set.
\paragraph{Experimental Details.}
Our implementation is based on the Composer package~\cite{mosaicml2022composer}. To ensure rigorous comparison, continued pre-training of TinyLLaMA v1 1.1B is carried out using identical data and experimental settings. We use a sequence length of 4096, a global train batch size of 64. For continued pre-training from the initial checkpoint (Stage 0), both APEX and the baseline adopt a learning rate of 1e-4. In training, we apply learning rate decay: 4e-5 in Stage 1, 1e-5 in Stage 2, and 5e-6 in Stage 3. At each stage, we select advantageous and disadvantageous dimensions based on evaluation statistics from the previous stage. Each experiment is run with different random seeds, and the results are averaged.
\paragraph{Evaluation.}
We follow the procedures established by Pythia~\cite{biderman2023pythiasuiteanalyzinglarge} and LLaMA2~\cite{touvron2023llama2openfoundation}. Specifically, we employ the lm-evaluation-harness~\cite{eval-harness} to evaluate our models. Specifically, we report zero-shot accuracy on ARC-Easy~\cite{clark2018think}, LAMBADA~\cite{paperno2016lambada}, LogiQA~\cite{liu2020logiqa}, PIQA~\cite{DBLP:conf/aaai/BiskZLGC20}, SciQ~\cite{welbl-etal-2017-crowdsourcing}, and WinoGrande~\cite{WinoGrande:conf/aaai/SakaguchiBBC20}. Additionally, we report the accuracy on tasks used by the OpenLLM Leaderboard~\cite{open-llm-leaderboard-v2}, including 10-shot HellaSwag~\cite{HellaSwag:conf/acl/ZellersHBFC19}, 25-shot ARC Challenge~\cite{clark2018think}, and 5-shot MMLU~\cite{hendrycks2021measuring}. To further assess the factual knowledge encoded in the model, we also report the accuracy on 32-shot BoolQ~\cite{clark2019boolq}.

We adhere to the license terms and conditions for all codes, models, and datasets used in this work. Specifically, we ensure that our use and distribution of these artifacts comply with their respective open-source licenses, and that our usage is consistent with their intended purpose, which is limited to research contexts where specified. For any artifacts we create, we clearly specify that they are intended for research use only, and verify that this intended use is compatible with the original access conditions, particularly for any datasets derived from sources restricted to research applications.

\begin{table}[t]
  \centering
\setlength{\tabcolsep}{15pt}
\begin{tabular}{@{}lc@{}}
\toprule
\textbf{Method} & \textbf{Exact Match}  \\
\midrule
Pre-trained & 12.0 \\
   \midrule
Mask Min 10\% & 12.0 ({-0.0}) \\
Mask Top 10\% & 0.0 ({-12.0}) \\
Mask Random 10\% & 11.5 ({-0.5}) \\
 \bottomrule
\end{tabular}
\caption{Comparison of the effects of masking advantageous and disadvantageous parameters on LLaMA2-7B for GSM8K. We mask 10\% of the parameters under each of the three settings.} 
\label{table-mask}
\end{table} 

\section{Extended Experiments}
\label{app-Exp}
\begin{table*}[t]
\centering
\setlength{\tabcolsep}{4pt}

        \begin{tabular}{cccccccc>{\columncolor{gray!20}}cc@{}}
            \toprule
\multicolumn{1}{c}{\multirow{2}{*}{\textbf{Method}}} & \multicolumn{1}{c}{\multirow{2}{*}{\textbf{\#Param.}}} & \textbf{MMLU} & \textbf{GSM8K} & \textbf{BBH} & \textbf{TyDiQA} & \textbf{TruthfulQA} & \textbf{HumanEval} &  \\
 &  & \scriptsize0-shot, EM & \scriptsize8-shot CoT, EM & \scriptsize3-shot CoT, EM & \scriptsize1-shot, F1 & \scriptsize0-shot, MC2 & \scriptsize0-shot, Pass@10 & \multirow{-2}{*}{\textbf{Avg.}} \\
            \midrule
            Full-FT & 100\% & 62.4 & 65.0 & 66.3 & 71.3 & 44.0 & 73.0 & 63.7 \\
            LoRA & 2.1\% & 61.0 & 64.0 & 63.4 & 70.9 & 45.8 & 67.8 & 62.2 \\
            DoRA & 2.1\% & 61.7 & 62.0 & 65.5 & 68.8 & 45.6 & 71.0 & 62.4 \\
            PiSSA & 2.1\% & 61.3 & 63.0 & 64.1 & 70.4 & 46.2 & 72.5 & 62.9 \\
            ReLoRA & 2.1\% & 61.0 & 65.0 & 60.7 & 70.5 & 45.3 & 68.0 & 61.8 \\
            S$^{2}$FT & 2.1\% & 63.1 & 61.0 & 64.4 & 70.9 & 43.8 & 71.3 & 62.4 \\
            APEX & 2.1\% & 62.4 & 64.5 & 68.7 & 71.8 & 44.4 & 73.1 & \textbf{64.2} \\
            \bottomrule
        \end{tabular}
\caption{Results for LLaMA3.1-8B. \#Param. denotes trainable parameters. }
\label{results-llama3}
\end{table*}
\begin{table*}[t]
\centering
\setlength{\tabcolsep}{4pt}
        \begin{tabular}{cccccccc>{\columncolor{gray!20}}cc@{}}
            \toprule
\multicolumn{1}{c}{\multirow{2}{*}{\textbf{Method}}} & \multicolumn{1}{c}{\multirow{2}{*}{\textbf{\#Param.}}} & \textbf{MMLU} & \textbf{GSM8K} & \textbf{BBH} & \textbf{TyDiQA} & \textbf{TruthfulQA} & \textbf{HumanEval} &  \\
 &  & \scriptsize0-shot, EM & \scriptsize8-shot CoT, EM & \scriptsize3-shot CoT, EM & \scriptsize1-shot, F1 & \scriptsize0-shot, MC2 & \scriptsize0-shot, Pass@10 & \multirow{-2}{*}{\textbf{Avg.}}\\
            \midrule
            Full-FT & 100\% & 53.4 & 41.5 & 47.8 & 59.0 & 44.4 & 41.5 & 47.9 \\
            LoRA & 1.9\% & 52.3 & 35.5 & 46.4 & 54.7 & 41.3 & 36.5 & 44.5 \\
            DoRA & 1.9\% & 51.5 & 40.5 & 44.4 & 56.5 & 43.8 & 35.2 & 45.3 \\
            PiSSA & 1.9\% & 53.7 & 32.5 & 50.8 & 58.1 & 44.6 & 34.5 & 45.7 \\
            ReLoRA & 1.9\% & 50.9 & 37.5 & 47.4 & 52.8 & 43.5 & 38.6 & 45.1 \\
            S$^{2}$FT & 1.9\% & 54.2 & 38.5 & 50.3 & 57.2 & 42.8 & 39.1 & 47.0 \\
            APEX & 1.9\% & 54.3 & 41.0 & 50.9 & 58.1 & 44.4 & 40.9 & \textbf{48.3} \\
            \bottomrule
        \end{tabular}
\caption{Results for LLaMA2-13B. \#Param. denotes trainable parameters. }
\label{results-llama2-13b}
\end{table*}

\subsection{Extended Observations}
\label{app-Exp:obs}
To further investigate whether parameter advantage can be reflected by activation magnitudes, we compare the effects of masking parameters with high activations to those with low activations. As shown in Table~\ref{table-mask}, masking high-activation 10\% parameters results in a significant decline in model performance, rendering the model nearly unusable. In contrast, masking low-activation 10\% parameters has little to no effect on performance. These results suggest that activation magnitudes can effectively distinguish advantageous parameters from disadvantageous ones, with the former playing a critical role in overall model performance.
\begin{table*}[t]
\centering
\setlength{\tabcolsep}{4pt}
        \begin{tabular}{cccccccc>{\columncolor{gray!20}}cc@{}}
            \toprule
\multicolumn{1}{c}{\multirow{2}{*}{\textbf{Method}}} & \multicolumn{1}{c}{\multirow{2}{*}{\textbf{\#Param.}}} & \textbf{MMLU} & \textbf{GSM8K} & \textbf{BBH} & \textbf{TyDiQA} & \textbf{TruthfulQA} & \textbf{HumanEval} &  \\
 &  & \scriptsize0-shot, EM & \scriptsize8-shot CoT, EM & \scriptsize3-shot CoT, EM & \scriptsize1-shot, F1 & \scriptsize0-shot, MC2 & \scriptsize0-shot, Pass@10 & \multirow{-2}{*}{\textbf{Avg.}}\\
            \midrule
            Full-FT & 100\% & 58.8 & 54.5 & 53.3 & 61.5 & 47.2 & 59.1 & 55.7 \\
            LoRA & 2.3\% & 56.3 & 47.5 & 54.5 & 58.2 & 45.8 & 53.2 & 52.6 \\
            DoRA & 2.3\% & 59.7 & 53.5 & 57.2 & 59.7 & 46.2 & 57.3 & 55.6 \\
            PiSSA & 2.3\% & 59.2 & 48.0 & 59.2 & 58.8 & 46.6 & 57.3 & 54.9 \\
            ReLoRA & 2.3\% & 59.2 & 51.0 & 55.6 & 59.7 & 46.2 & 56.7 & 54.7 \\
            S$^{2}$FT & 2.3\% & 60.1 & 54.5 & 58.9 & 58.8 & 45.2 & 58.6 & 56.0 \\
            APEX & 2.3\% & 59.8 & 54.0 & 59.0 & 61.9 & 45.9 & 62.7 & \textbf{57.2} \\
            \bottomrule
        \end{tabular}
\caption{Results for Mistral-7B-v0.1. \#Param. denotes trainable parameters. }
\label{results-mistral}
\end{table*}
\subsection{Experiments on Additional Models}
\label{app-Exp:models}
In Tables~\ref{results-llama3}--\ref{results-mistral}, we report the results of instruction tuning experiments using LLaMA3.1-8B, LLaMA2-13B, and Mistral-7B-v0.1 as base models. As shown, similar to the performance trends observed with fine-tuning on LLaMA2-7B in Table~\ref{results-it}, APEX outperforms all parameter-efficient fine-tuning baselines in terms of average performance and achieves leading results on the majority of evaluation tasks. Notably, APEX attains performance comparable to full-parameter fine-tuning while only requiring around 2\% of the trainable parameters. These results demonstrate that advantage expansion is effective across different base models and can consistently improve training performance.

\begin{figure}[t]
    \subfloat[Number of parameters introduced by the expansion operator.]{ \includegraphics[width=0.48\linewidth]{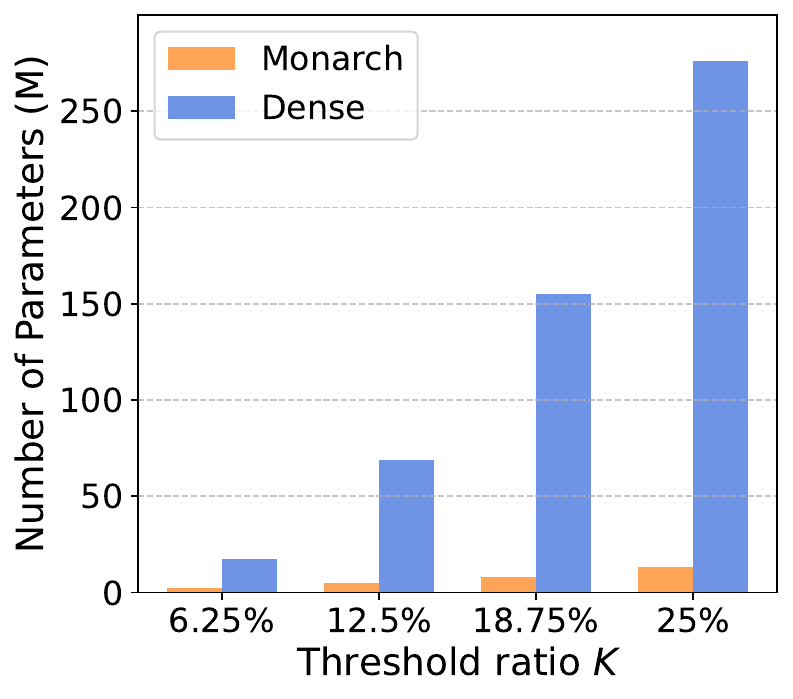}\label{fig:efficiency-param}}
    \hfill
    \subfloat[Training time of different methods on LLaMA2-7B.]{	\includegraphics[width=0.46\linewidth]{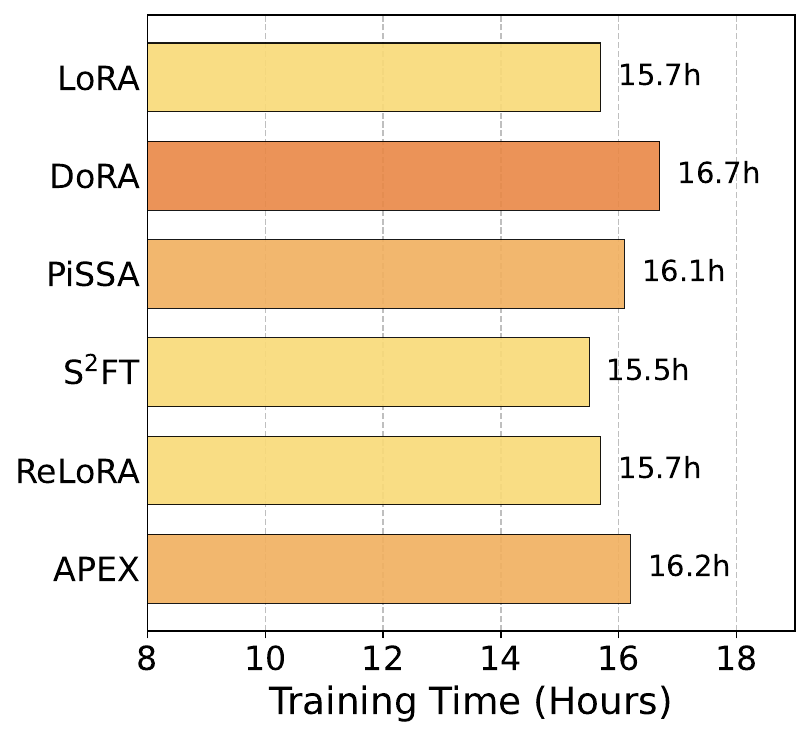}\label{fig:efficiency-time}}
\caption{Efficiency analysis of APEX.}
\end{figure}
\subsection{Efficiency Analysis}
\label{app-Exp:efficiency}
\noindent\textbf{Efficiency Analysis of the Expansion Operator.} In Figure~\ref{fig:efficiency-param}, we compare the trend of parameters introduced by the expansion operator when implemented with Dense matrices versus Monarch matrices under varying threshold ratios ($K$). As $K$ increases from 6.25\% to 25\%, the parameter count for the Dense implementation rises sharply, from 17.2M to 275M. In contrast, the Monarch-based expansion operator consistently maintains a much lower parameter scale, significantly below that of the Dense approach. This indicates that Monarch decomposition can effectively compress the parameter overhead of the expansion operator, enabling APEX to achieve parameter-efficient expansion with minimal additional parameters. Consequently, APEX supports the expansion of more advantageous parameters without compromising the lightweight nature of PEFT.

\noindent\textbf{Training Efficiency Analysis.} Figure~\ref{fig:efficiency-time} presents a comparison of end-to-end training time across different methods under identical model and data settings. The results show that APEX requires approximately 16.2 hours of training, only about 0.5 hours longer than LoRA, on par with PiSSA, and faster than DoRA. Thanks to the integration of advantage parameter evaluation in the forward pass and the lightweight design of the expansion operator, the evaluation and expansion steps introduced by APEX do not noticeably slow down training throughput. Overall, APEX delivers stable performance improvements while keeping both the additional parameter and actual training time overheads at the same order of magnitude as mainstream PEFT methods, demonstrating its practical feasibility and efficiency advantage.

\begin{table}[t]
  \centering
\setlength{\tabcolsep}{18pt}
\begin{tabular}{@{}lc@{}}
\toprule
\textbf{Method} & \textbf{Average}  \\
\midrule
HFT-52\% & 42.9 \\
   \midrule
 APEX-52\% &  \\
\ \ \ \textit{w. Act. Rand.} & 43.2 (+0.3) \\
\ \ \ \textit{w. Act. Avg.} & 43.7 (+0.8) \\
\ \ \ \textit{w. Act. Rank.} & 44.0 (+1.1) \\
\ \ \ \textit{w. Grad. Rank.} & 44.1 (+1.2) \\
 \bottomrule
\end{tabular}
\caption{Analysis of assessment method in APEX.} 
\label{table-assessment}
\end{table} 

\subsection{Extended Method Analysis}
\label{app-Exp:analysis}
\noindent\textbf{Comparison of Different Assessment Methods.} We compare the performance of four assessment methods: no assessment (i.e., \textit{Act. Rand}.), using the average activations (i.e., \textit{Act. Avg.}), our proposed metric based on relative activation rankings (i.e., \textit{Act. Rank}), and gradient-based ranking (i.e., \textit{Grad. Rank}). As shown in Table~\ref{table-assessment}, employing the average activations as an assessment method improved the average performance on instruction tuning tasks by 0.5 points compared to no assessment, demonstrating the effectiveness of activations in selecting advantageous parameters. Furthermore, when incorporating our proposed relative activation ranking as the assessment metric, the performance improved further. This indicates that the relative ranking of activations can effectively mitigate the fluctuations caused by data variability, leading to more accurate assessment results. Crucially, \textit{Act. Rank} achieves performance comparable to \textit{Grad. Rank} while avoiding the substantial overhead of gradient-based assessment, which requires backpropagation
and storing gradients for all parameters. In contrast, \emph{Act. Rank} only records forward-pass
activations, providing a lightweight assessment for advantageous parameters.
\begin{figure}[t]
    \centering
\includegraphics[width=0.8\columnwidth]{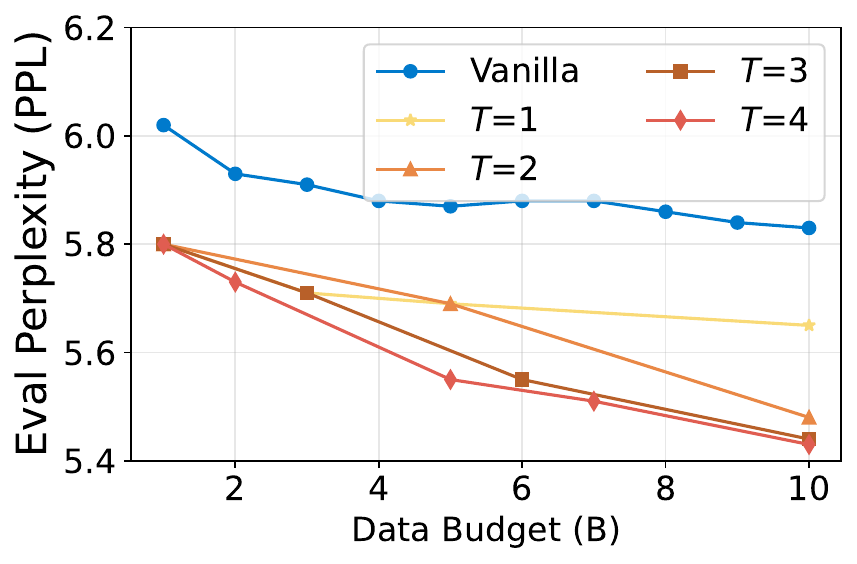}
\caption{Perplexity curves during continued pre-training with varying stage counts $T$.}
\label{fig:cpt-impact-stage}
\end{figure}
\begin{table*}[t]
\centering
\setlength{\tabcolsep}{3.5pt}
        \begin{tabular}{cccccccccc@{}}
            \toprule
\multicolumn{1}{c}{\multirow{2}{*}{\textbf{Method}}} & \multicolumn{1}{c}{\multirow{2}{*}{\textbf{\#Param.}}} & \textbf{MMLU} & \textbf{GSM8K} & \textbf{BBH} & \textbf{TyDiQA} & \textbf{TruthfulQA} & \textbf{HumanEval} & \multirow{2}{*}{\textbf{Avg.}} \\
 &  & \scriptsize0-shot, EM & \scriptsize8-shot CoT, EM & \scriptsize3-shot CoT, EM & \scriptsize1-shot, F1 & \scriptsize0-shot, MC2 & \scriptsize0-shot, Pass@10 & \\
            \midrule
            Pre-trained$^{\dagger}$ & - & 41.6 & 12.0 & 39.9 & 48.4 & 38.5 & 26.2 & 34.4 \\
            S$^{2}$FT & 2.4\% & 47.6 & 25.0 & 45.5 & 52.8 & 43.7 & 33.9 & 41.4 \\
            Act-Regu & 2.4\% & {23.8} & 30.5 & 41.2 & 51.8 & {0.0} & 31.1 & {29.7} \\
            \rowcolor{gray!20}APEX & 2.4\% & 48.1 & 28.5 & 45.7 & 54.5 & 45.5 & 33.2 & 42.6 \\
            \bottomrule
        \end{tabular}
\caption{Results of using activation regularization (Act-Regu) for LLaMA2-7B. \#Param. denotes trainable parameters. All results with $^{\dagger}$ are taken from HFT~\cite{hui-etal-2025-hft}. }
\label{results-regu}
\end{table*}

\noindent \textbf{Effect of the number of stages $T$ for CPT settings.} Unlike instruction tuning, which typically involves multi-epoch training, continued pre-training often follows a single-pass paradigm over a massive corpus. Figure~\ref{fig:cpt-impact-stage} shows the eval PPL results of APEX with 1, 2, 3, and 4 stages under a 10B continued pre-training budget. First, APEX consistently and significantly outperforms traditional methods across all stage settings, as evidenced by a more stable and pronounced decrease in PPL. Second, increasing the number of stages can further enhance the performance of APEX. For example, $T=3$, which has more stages than $T=2$ under the same training data, is able to better leverage the benefits of APEX in each stage and thus achieves a lower eval PPL. Finally, the performance improvement from adding more APEX stages has its limits; as the number of stages increases, the difference between the model’s advantages and disadvantages narrows, leading to diminishing returns. As a result, the gap in final evaluation performance between $T=3$ and $T=4$ is relatively small. Based on this trade-off between performance gains and training complexity, we adopt $T=3$ as the default setting for all subsequent continued pre-training experiments.
\begin{table*}[h]
\centering
\begin{tabular}{lcccc}
\toprule
 & \textbf{MHA Top-K} & \textbf{MHA Min-K} & \textbf{FFN Top-K} & \textbf{FFN Min-K} \\
\midrule
Layer 5  & 0.75 & 0.60 & 0.78 & 0.63 \\
Layer 15 & 0.72 & 0.47 & 0.81 & 0.59 \\
Layer 25 & 0.69 & 0.54 & 0.87 & 0.66 \\
\midrule
Overall Avg. & 0.72 & 0.54 & 0.82 & 0.63 \\
\bottomrule
\end{tabular}
\caption{Jaccard similarity of Top-K and Min-K parameter sets between consecutive APEX stages on LLaMA2-7B ($T=2$).}
\label{tab:jaccard}
\end{table*}
\paragraph{Discussion of the Relationship Between Activation Distributions and Model Performance.}
\label{app-Exp:further}

In Section~\ref{sec-obs}, we observed that strong models exhibit more uniform activation distributions. This raises a critical question: Can we improve performance by explicitly enforcing this statistical uniformity? To answer this, we contrast two fundamentally different paradigms.

\textbf{Activation Regularization (Act-Regu).} 
A natural hypothesis is that directly enforcing uniform activation distributions through regularization would improve performance. To test this, we implement Act-Regu, which penalizes the standard deviation of activations during training. However, as shown in Table~\ref{results-regu}, Act-Regu substantially degrades performance, reducing the average score from 42.6 to 29.7 and causing TruthfulQA accuracy to fall to 0\%. This result indicates that the more uniform activation distributions observed in strong models reflect effective parameter utilization rather than provide a direct means of improving it. Forcibly suppressing high activations destroys the informative representations that these parameters have learned.

\textbf{APEX.} 
Rather than suppressing advantageous parameters, APEX leverages them to guide the optimization of disadvantageous ones. As analyzed in Appendix~\ref{method-theory} (Lemma~\ref{lemma1}), the expansion operator introduces a preconditioned correction term $({\mathbf{W}^P}{\mathbf{W}^P}^\top)\hat{\mathbf{G}}$ that projects gradient updates onto the subspace spanned by advantageous parameters. This subspace, having been shaped by reliable gradient signals during prior training, provides a higher signal-to-noise ratio for optimization. Consequently, disadvantageous parameters receive more stable gradient guidance without overwriting the learned representations in advantageous ones.

The contrast between these approaches highlights a fundamental distinction. Act-Regu treats activation uniformity as a target, destroying useful information to achieve it. APEX instead treats advantageous parameters as a resource, using their learned structure to improve optimization in underutilized subspaces. The flatter activation distribution observed after APEX training (Figure~\ref{fig:apex_vs_vanilla}) emerges naturally as a byproduct of broader parameter utilization, not as an imposed constraint. This validates our hypothesis that the path to enhanced performance lies in expanding advantageous capacity rather than suppressing it.

\subsection{Stability of Stage-wise Parameter Re-selection}
\label{app:stability}
Since APEX re-selects advantageous and disadvantageous parameter sets at each training stage, a natural question is whether these sets remain stable across stages. We conduct an analysis on the LLaMA2-7B instruction tuning setting with $T=2$ stages. We extract the Top-K and Min-K parameter sets for each stage and compute the Jaccard similarity $J = |S_t \cap S_{t+1}| / |S_t \cup S_{t+1}|$ between consecutive stages. Table~\ref{tab:jaccard} reports the results for representative layers.
 
As shown in Table~\ref{tab:jaccard}, the Top-K (advantageous) sets maintain high consistency across consecutive stages, indicating that core advantageous parameters do not fluctuate drastically due to stage transitions and that the assessment mechanism is reliable. Meanwhile, the Jaccard similarity of the Min-K (disadvantageous) sets is systematically lower than that of Top-K, with gaps of 0.18 in MHA and 0.19 in FFN. This asymmetry suggests that some parameters originally classified as disadvantageous exit the Min-K set after expansion training, which is consistent with the intended effect of APEX. We also note that FFN modules exhibit higher Jaccard similarity than MHA modules across all layers, which is consistent with Figure~\ref{pilot1}, where FFN activation distributions are sharper, implying clearer Top/Min boundaries and naturally more stable cross-stage selection.
\subsection{Representational Diversity After Expansion}
\label{app:diversity}
To verify that the expansion mechanism preserves representational diversity across attention heads, we extract the fine-tuned LLaMA2-7B model weights and compute the average pairwise cosine similarity between output states of different attention head weight matrices within the same layer. Specifically, for each layer, we compute $\frac{1}{H(H-1)} \sum_{i \neq j} \cos(\mathbf{W}_O^{[i,:]}, \mathbf{W}_O^{[j,:]})$, where $H$ is the number of heads.
 
\begin{table}[t]
\centering
\setlength{\tabcolsep}{3pt}
\begin{tabular}{lcc}
\toprule
 & \textbf{Full-FT} & \textbf{APEX} \\
\midrule
Layer 5  & 0.001767 & 0.002452 \\
Layer 15 & 0.001347 & 0.001063 \\
Layer 25 & 0.001525 & 0.002292 \\
\midrule
Overall Avg. & 0.001546 & 0.001936 \\
\bottomrule
\end{tabular}
\caption{Average pairwise cosine similarity between attention head output projections on LLaMA2-7B.}
\label{tab:cosine_diversity}
\end{table}
 
As shown in Table~\ref{tab:cosine_diversity}, cosine similarity values remain at the very low $10^{-3}$ level for both Full-FT and APEX, indicating that the learned representations of different heads remain nearly orthogonal. The overall average similarity after APEX training is comparable to that of Full-FT, providing empirical evidence that APEX preserves representational diversity and does not induce representation collapse.
 
This result aligns with the design principle described in Section~\ref{method-operator}. Since the Monarch-parameterized operator is zero-initialized, it functions as a structured preconditioner during gradient updates, as shown in Lemma~\ref{lemma1}. Rather than forcing disadvantageous parameters to become copies of advantageous ones, APEX guides them toward the advantageous subspace while retaining their capacity to learn independent, orthogonal patterns via standard gradient updates. Furthermore, the multi-stage re-selection mechanism allows these expansion directions to continually adapt across different training stages.
\subsection{Robustness Across Random Seeds and Calibration Data}
\label{app:perseed}
All experimental results reported in the main paper are averaged over three independent runs with different random seeds (0, 42, 4242). Importantly, in our implementation, the data subset $\mathcal{D}_\text{subset}$ used for pre-assessment in the first stage is determined by the random seed and sampled from the training set. This means that the three runs use different first-stage calibration data.
\begin{table}[t]
\centering
\setlength{\tabcolsep}{3pt}
\begin{tabular}{lcccc}
\toprule
\textbf{Method} & \textbf{Seed 0} & \textbf{Seed 42} & \textbf{Seed 4242} & \textbf{AVG} \\
\midrule
Full-FT & 42.8 & 43.2 & 43.1 & 43.1 \\
APEX-52\% & 44.3 & 44.0 & 43.6 & 44.0 \\
\bottomrule
\end{tabular}
\caption{Per-seed average performance for instruction tuning on LLaMA2-7B.}
\label{tab:perseed_sft}
\end{table}

Table~\ref{tab:perseed_sft} reports the per-seed average performance for instruction tuning on LLaMA2-7B. The per-seed results reveal that the improvement of APEX over baselines is consistent, as even the worst single run of APEX in instruction tuning exceeds the best Full-FT run. Moreover, despite using different initial calibration subsets across the three runs, the final performance remains highly stable. This robustness stems from two aspects of our design. First, the calibration subset is randomly sampled from a diverse, multi-domain training corpus, so the identified advantageous parameters reflect the model's fundamental representational strengths rather than domain-specific biases. Second, APEX’s multi-stage mechanism acts as a self-correcting process. Beyond the first stage, advantage assessment is based on activation statistics accumulated during training on the full data distribution, and the influence of the initial calibration gradually diminishes as training progresses.
\begin{table}[t]
\centering
\small
\setlength{\tabcolsep}{3pt}
\begin{tabular}{lccccccc}
\toprule
\textbf{Method} & \textbf{MML.} & \textbf{GSM.} & \textbf{BBH} & \textbf{TyD.} & \textbf{Tru.} & \textbf{Hum.} & \textbf{Avg.}  \\
\midrule
Full-FT & 25.7 & 3.5 & 23.6 & 24.2 & 38.5 & 7.8 & 20.6 \\
APEX & 25.8 & 3.5 & 25.5 & 24.5 & 38.6 & 9.2 & 21.2 \\
\bottomrule
\end{tabular}
\caption{Instruction tuning results for Sheared-LLaMA-1.3B.}
\label{tab:small_model}
\end{table}
\subsection{Instruction Tuning on Small Models}
\label{app:small_model}
To further validate that APEX helps small models make better use of their parameters, we conduct experiments on Sheared-LLaMA-1.3B~\citep{xia2024sheared}. As shown in Table~\ref{tab:small_model}, APEX achieves consistent improvements over Full-FT. Combined with the continued pre-training results on TinyLLaMA-1.1B (Table~\ref{table:pt}), these experiments confirm that APEX is effective on models ranging from 1.1B to 13B.
\section{Theoretical Perspective: Gradient Signal Alignment}
\label{method-theory}
We analyze why expanding \emph{advantageous} parameters can improve optimization from a \emph{gradient signal alignment} perspective. The key point in our actual training setup is that the disadvantageous block $\mathbf{W}_N$ is still updated, while the expansion matrix $\mathbf{M}$ is \emph{zero-initialized}. Consequently, APEX does \emph{not} forcibly fuse $\mathbf{W}_P$ into $\mathbf{W}_N$; instead, it provides an additional, adaptive correction term that becomes active only when the residual gradient has a non-trivial projection onto the advantageous feature span.
\subsection{Setup}

Consider a single linear layer with weights $\mathbf{W}\in\mathbb{R}^{d_{\text{in}}\times d_{\text{out}}}$ and input $\mathbf{x}\in\mathbb{R}^{1 \times d_{\text{in}}}$. 
In our proposed method, the columns of $\mathbf{W}$ correspond to computational units (e.g., neurons in FFN or heads in Attention). We partition these columns into two sets based on their output activation magnitudes assessed on the dataset.
Let $\mathcal{I}_P$ and $\mathcal{I}_N$ denote the index sets for advantageous and disadvantageous parameters, respectively. We conceptually partition the weight matrix as:
\begin{equation}
\mathbf{W} = \bigl[\mathbf{W}_P,\ \mathbf{W}_N\bigr],
\end{equation}
where $\mathbf{W}_P \in \mathbb{R}^{d_{\text{in}} \times k}$ corresponds to the subset of parameters generating high-magnitude activations (advantageous), and $\mathbf{W}_N \in \mathbb{R}^{d_{\text{in}} \times (d_{\text{out}}-k)}$ corresponds to those with low-magnitude activations (disadvantageous). 
Accordingly, the layer output can be formulated as $\mathbf{y} = \mathrm{Concat}(\mathbf{y}_P, \mathbf{y}_N),$ where $\mathbf{y}_P = \mathbf{x}\mathbf{W}_P$ and $\mathbf{y}_N = \mathbf{x}\mathbf{W}_N$.

\noindent\textbf{Gradient Signal Quality.}
To analyze the optimization dynamics, we model the stochastic gradient $\widehat{\mathbf{G}}$ (computed on a mini-batch) as the true population gradient $\mathbf{G}_{\star}$ perturbed by zero-mean noise $\mathbf{\Xi}$:
\begin{equation}
\widehat{\mathbf{G}} = \mathbf{G}_{\star} + \mathbf{\Xi}, \quad \text{where } \mathbb{E}[\mathbf{\Xi}] = \mathbf{0}.
\label{eq:sgd_decomp}
\end{equation}
We define the Signal-to-Noise Ratio (SNR) as
\begin{equation}
\mathrm{SNR}(\widehat{\mathbf{G}}) \ :=\ \frac{\|\mathbf{G}_{\star}\|_F^2}{\mathbb{E}[\|\mathbf{\Xi}\|_F^2]}.
\label{eq:snr_def}
\end{equation}

\noindent\textbf{Assumption 1 (Activation-selected subspace has higher SNR).}
\textit{We assume that the column space of $\mathbf{W}_P$ selected by high activations captures a larger fraction of the useful gradient signal or yields more stable gradient estimates across mini-batches.}

Formally, let $\mathbf{P}$ denote the orthogonal projector onto $\mathrm{col}(\mathbf{W}_P)$:
\begin{equation}
\mathbf{P}\ :=\ \mathbf{W}_P(\mathbf{W}_P^\top \mathbf{W}_P)^{\dagger}\mathbf{W}_P^\top,
\label{eq:proj_def}
\end{equation}
where $(\cdot)^{\dagger}$ is the pseudoinverse. We assume that the gradient component aligned with $\mathrm{col}(\mathbf{W}_P)$ has a higher effective SNR than the orthogonal component:
\begin{equation}
\begin{aligned}
\mathrm{SNR}(\mathbf{P}\widehat{\mathbf{G}})\ \ge\ \alpha\cdot 
&\mathrm{SNR}\bigl((\mathbf{I}-\mathbf{P})\widehat{\mathbf{G}}\bigr),\\
\quad &\text{for some } \alpha>1. 
\end{aligned}
\label{eq:assump_subspace_snr}
\end{equation}
Here $\alpha>1$ quantifies the multiplicative SNR advantage of the gradient component aligned with
$\mathrm{col}(\mathbf{W}_P)$ over its orthogonal component. This assumption is intentionally coarse and does not claim that every low-activation unit must have low-SNR gradients.

\noindent\textbf{Justification via Empirical Observations.}
Our masking experiments show that removing low-activation units ($\mathbf{W}_N$) causes negligible performance degradation, suggesting that these units are weakly coupled to the current function implemented by the layer. In such a regime, updates that rely on directions orthogonal to the active feature subspace may be inefficient on average, whereas reusing the already-active feature subspace $\mathrm{col}(\mathbf{W}_P)$ provides a structured way to express corrections. This motivates us to adopt Assumption~\eqref{eq:assump_subspace_snr} as a coarse, subspace-level abstraction: the population gradient contains a more meaningful component aligned with $\mathrm{col}(\mathbf{W}_P)$ than with its orthogonal complement. Importantly, APEX does not force fusion: $\mathbf{M}_0=\mathbf{0}$ preserves the initial function, and the $\mathbf{M}$-path is only driven by the aligned component through $\nabla_{\mathbf{M}}\mathcal{L}=\mathbf{W}_P^\top \widehat{\mathbf{G}}$.

\subsection{Baseline Gradients and the Optimization Dilemma}
Consider optimizing the disadvantageous block $\mathbf{W}_N$ under standard SGD:
\begin{equation}
\mathbf{W}_N \leftarrow \mathbf{W}_N - \eta\,(\mathbf{G}_{N,\star} + \mathbf{\Xi}_N).
\end{equation}
Under Assumption~\eqref{eq:assump_subspace_snr}, gradient components that are \emph{not} aligned with $\mathrm{col}(\mathbf{W}_P)$ tend to be less reliable on average, so naive updates may spend many steps in variance-dominated directions. This motivates \emph{gradient guidance}: leveraging the activation-selected subspace $\mathrm{col}(\mathbf{W}_P)$ to supply a structured correction path while still allowing $\mathbf{W}_N$ to update directly when its gradients are informative.

\subsection{APEX Reparameterization with Concurrent Updates}

APEX reparameterizes the disadvantageous block as
\begin{equation}
\begin{aligned}
    \mathbf{W}_N' \ &:=\ \mathbf{W}_N + \mathbf{W}_P\mathbf{M},
\\
&\text{where}\ \mathbf{M}\in\mathbb{R}^{k\times(d_{\text{out}}-k)},\ \mathbf{M}_0=\mathbf{0}.
\end{aligned}
\label{eq:apex_param}
\end{equation}
Consequently, the output of the disadvantageous units becomes:$\mathbf{y}_N' = \mathbf{x}\mathbf{W}_N' = \mathbf{x}\mathbf{W}_N + \mathbf{x}\mathbf{W}_P\mathbf{M}$.
Importantly, in our training, \textbf{$\mathbf{W}_N$ is still updated} together with $\mathbf{M}$ (and typically $\mathbf{W}_P$ as well).

Let $\mathbf{G} := \nabla_{\mathbf{W}_N'}\mathcal{L} \in \mathbb{R}^{d_{\text{in}}\times(d_{\text{out}}-k)}$ denote the gradient w.r.t.\ the \emph{effective} block $\mathbf{W}_N'$.
By the chain rule,
\begin{equation}
\nabla_{\mathbf{W}_N}\mathcal{L} = \mathbf{G},
\qquad
\nabla_{\mathbf{M}}\mathcal{L} = \mathbf{W}_P^\top \mathbf{G}.
\label{eq:chain_rule}
\end{equation}

\begin{lemma}
\label{lemma1}
\textnormal{(Effective update equals vanilla step plus preconditioned correction.)}
Consider SGD updates with learning rates $\eta_N,\eta_M>0$:
\begin{equation}
\begin{aligned}
    &\mathbf{W}_N \leftarrow \mathbf{W}_N - \eta_N\,\widehat{\mathbf{G}},
\\ &\mathbf{M} \leftarrow \mathbf{M} - \eta_M\,\mathbf{W}_P^\top \widehat{\mathbf{G}}.
\end{aligned}
\label{eq:sgd_updates}
\end{equation}
At initialization, where $\mathbf{M}_0=\mathbf{0}$, the induced update to $\mathbf{W}_N'$ is approximated to first order as
\begin{equation}
\Delta \mathbf{W}_N'
\ \approx\
-\eta_N\,\widehat{\mathbf{G}}
\ -\
\eta_M\,(\mathbf{W}_P\mathbf{W}_P^\top)\widehat{\mathbf{G}}.
\label{eq:effective_update}
\end{equation}
\end{lemma}
\paragraph{Proof of Lemma~1.}
Recall the reparameterization $\mathbf{W}_N'=\mathbf{W}_N+\mathbf{W}_P\mathbf{M}$.
For one SGD step, write the increments as
\begin{equation}
\begin{aligned}
        &\Delta \mathbf{W}_N := \mathbf{W}_N^{+}-\mathbf{W}_N,\\
&\Delta \mathbf{M} := \mathbf{M}^{+}-\mathbf{M},\\
&\Delta \mathbf{W}_P := \mathbf{W}_P^{+}-\mathbf{W}_P.
\end{aligned}
\end{equation}
Then by direct expansion,
\begin{equation}
\begin{aligned}
\Delta \mathbf{W}_N'
&= \mathbf{W}_N^{+} + \mathbf{W}_P^{+}\mathbf{M}^{+} - (\mathbf{W}_N + \mathbf{W}_P\mathbf{M}) \notag\\
&= \Delta \mathbf{W}_N + (\mathbf{W}_P+\Delta \mathbf{W}_P)(\mathbf{M}+\Delta \mathbf{M})\\&\qquad-\mathbf{W}_P\mathbf{M} \notag\\
&= \Delta \mathbf{W}_N + \mathbf{W}_P\Delta \mathbf{M} + (\Delta \mathbf{W}_P)\mathbf{M}\\&\qquad + (\Delta \mathbf{W}_P)(\Delta \mathbf{M}). \label{eq:delta_wnp_expand}
\end{aligned}
\end{equation}

Using the chain rule $\nabla_{\mathbf{W}_N}\mathcal{L}=\mathbf{G}$ and
$\nabla_{\mathbf{M}}\mathcal{L}=\mathbf{W}_P^\top \mathbf{G}$, the SGD updates
\begin{equation}
    \mathbf{W}_N^{+}=\mathbf{W}_N-\eta_N \widehat{\mathbf{G}},\ \ 
\mathbf{M}^{+}=\mathbf{M}-\eta_M \mathbf{W}_P^\top \widehat{\mathbf{G}}
\end{equation}
can imply:
\begin{equation}
\Delta \mathbf{W}_N = -\eta_N \widehat{\mathbf{G}},\ \ 
\Delta \mathbf{M} = -\eta_M \mathbf{W}_P^\top \widehat{\mathbf{G}}.
\end{equation}
Substituting into \eqref{eq:delta_wnp_expand} gives
\begin{equation}
\begin{aligned}
   \Delta \mathbf{W}_N'
&=
-\eta_N \widehat{\mathbf{G}}
-\eta_M(\mathbf{W}_P\mathbf{W}_P^\top)\widehat{\mathbf{G}}
\\&+(\Delta \mathbf{W}_P)\mathbf{M}
+(\Delta \mathbf{W}_P)(\Delta \mathbf{M}).  
\end{aligned}
\label{eq:delta_full_with_wp}
\end{equation}
At initialization $\mathbf{M}_0=\mathbf{0}$, the term $(\Delta \mathbf{W}_P)\mathbf{M}$ is exactly zero on the first step.
More generally during training, if $\|\Delta \mathbf{W}_P\|=O(\eta_P)$ and $\|\Delta \mathbf{M}\|=O(\eta_M)$, then
$(\Delta \mathbf{W}_P)(\Delta \mathbf{M})=O(\eta_P\eta_M)$ is a higher-order term.
Neglecting these higher-order contributions yields the first-order approximation
\begin{equation}
\Delta \mathbf{W}_N'
\approx
-\eta_N \widehat{\mathbf{G}}
-\eta_M(\mathbf{W}_P\mathbf{W}_P^\top)\widehat{\mathbf{G}},
\end{equation}

\paragraph{Interpretation.}
Equation \eqref{eq:effective_update} shows that APEX does \textbf{not} replace the vanilla update of $\mathbf{W}_N$; it adds an extra correction term that left-multiplies the gradient by $\mathbf{W}_P\mathbf{W}_P^\top$, i.e., a \emph{subspace-aligned preconditioner}. When $\mathbf{W}_P$ is reasonably conditioned, one may view $\mathbf{W}_P\mathbf{W}_P^\top$ as a scaled projector toward $\mathrm{col}(\mathbf{W}_P)$, steering part of the update to lie in the span of already-learned strong features.

\subsection{Why Zero-Initialized $\mathbf{M}$ Does Not Force Fusion: Adaptive Activation}

Since $\mathbf{M}_0=\mathbf{0}$, the initial function is unchanged. Moreover, the update magnitude of $\mathbf{M}$ is governed by the projected gradient:
\begin{equation}
\|\Delta \mathbf{M}\|_F
=
\eta_M \|\mathbf{W}_P^\top \widehat{\mathbf{G}}\|_F.
\label{eq:m_update_gate}
\end{equation}
Thus, if $\mathbf{W}_N$ is already sufficient such that the residual gradient $\widehat{\mathbf{G}}$ becomes small, or if $\widehat{\mathbf{G}}$ is (approximately) orthogonal to $\mathrm{col}(\mathbf{W}_P)$ so that $\mathbf{W}_P^\top \widehat{\mathbf{G}}\approx \mathbf{0}$, then $\mathbf{M}$ remains near zero. In this sense, APEX behaves as a \emph{data-adaptive correction path}: it activates only when there exists a residual component that can be efficiently reduced through the advantageous feature span.

\subsection{Variance Reduction and SNR Improvement of the APEX Correction}
We now formalize how the additional correction term in Eq.~\eqref{eq:effective_update} redirects part of the update from lower-SNR components to higher-SNR components, using a bias--variance decomposition. 
Following Eq.~\eqref{eq:sgd_decomp}, write the stochastic gradient as $\widehat{\mathbf{G}}=\mathbf{G}_\star+\mathbf{\Xi}$. 
Let $\mathbf{P}$ be the orthogonal projector onto $\mathrm{col}(\mathbf{W}_P)$ as defined in Eq.~\eqref{eq:proj_def}.

\begin{lemma}
\label{lemma2}
\textnormal{(Orthogonal Projection Does Not Increase Gradient Noise Energy.)}
\textit{For any zero-mean noise matrix $\mathbf{\Xi}$,}
\begin{equation}
\mathbb{E}\bigl[\|\mathbf{P}\mathbf{\Xi}\|_F^2\bigr]\ \le\ \mathbb{E}\bigl[\|\mathbf{\Xi}\|_F^2\bigr].
\label{eq:noise_reduce}
\end{equation}
\end{lemma}
Define the projected signal and noise scales:
\begin{equation}
S_\parallel := \|\mathbf{P}\mathbf{G}_\star\|_F^2,\ \ 
N_\parallel := \mathbb{E}\bigl[\|\mathbf{P}\mathbf{\Xi}\|_F^2\bigr],
\end{equation}
and the corresponding SNR:
\begin{equation}
\mathrm{SNR}_\parallel\ :=\ \frac{S_\parallel}{N_\parallel}.
\label{eq:snr_proj}
\end{equation}
Lemma~2 implies $N_\parallel\le \mathbb{E}\|\mathbf{\Xi}\|_F^2$, i.e., the \emph{orthogonally projected} path which captures pure subspace alignment to $\mathrm{col}(\mathbf{W}_P)$ is no more noisy in energy than the original noise.
In APEX, the actual correction uses $\mathbf{W}_P\mathbf{W}_P^\top$, which additionally applies magnitude scaling within $\mathrm{col}(\mathbf{W}_P)$; this scaling is controlled by $\eta_M$.
Whether it helps depends on how much signal is retained in $\mathbf{P}\mathbf{G}_\star$, which is precisely what Eq.\eqref{eq:m_update_gate} adaptively measures during training.

\paragraph{Proof of Lemma~2.}
Let $\mathbf{P}$ be an orthogonal projector onto $\mathrm{col}(\mathbf{W}_P)$, i.e.,
$\mathbf{P}^\top=\mathbf{P}$ and $\mathbf{P}^2=\mathbf{P}$, hence $0\preceq \mathbf{P}\preceq \mathbf{I}$.
For any random matrix $\mathbf{\Xi}$ with finite second moment,
\begin{equation}
\begin{aligned}
        \mathbb{E}\bigl[\|\mathbf{P}\mathbf{\Xi}\|_F^2\bigr]
&= \mathbb{E}\bigl[\mathrm{Tr}((\mathbf{P}\mathbf{\Xi})^\top(\mathbf{P}\mathbf{\Xi}))\bigr]\\
&= \mathbb{E}\bigl[\mathrm{Tr}(\mathbf{\Xi}^\top \mathbf{P}^\top \mathbf{P}\mathbf{\Xi})\bigr] \\
&= \mathbb{E}\bigl[\mathrm{Tr}(\mathbf{\Xi}^\top \mathbf{P}\mathbf{\Xi})\bigr]\\
&= \mathbb{E}\bigl[\mathrm{Tr}(\mathbf{\Xi}\mathbf{\Xi}^\top \mathbf{P})\bigr]. 
\end{aligned}
\label{eq:proj_noise_trace}
\end{equation}
Since $\mathbf{\Xi}\mathbf{\Xi}^\top \succeq 0$ and $0\preceq \mathbf{P}\preceq \mathbf{I}$, we have
\begin{equation}
\mathrm{Tr}(\mathbf{\Xi}\mathbf{\Xi}^\top \mathbf{P})
\le
\mathrm{Tr}(\mathbf{\Xi}\mathbf{\Xi}^\top \mathbf{I})
=
\mathrm{Tr}(\mathbf{\Xi}\mathbf{\Xi}^\top)
=
\|\mathbf{\Xi}\|_F^2.
\end{equation}
Taking expectation on both sides and combining with Eq.\eqref{eq:proj_noise_trace} yields
\begin{equation}
\mathbb{E}\bigl[\|\mathbf{P}\mathbf{\Xi}\|_F^2\bigr]\le \mathbb{E}\bigl[\|\mathbf{\Xi}\|_F^2\bigr],
\end{equation}

\subsection{Summary of Analysis}
With concurrent updates of $\mathbf{W}_N$ and zero-initialized $\mathbf{M}$, APEX is best viewed as an \emph{auxiliary preconditioned correction path}:
\begin{itemize}
\item \textbf{No forced fusion:} $\mathbf{M}_0=\mathbf{0}$ preserves the initial function, and the $\mathbf{M}$-path is activated only when the residual gradient has a non-trivial projection onto the advantageous span, i.e., when $\|\mathbf{W}_P^\top \widehat{\mathbf{G}}\|$ is non-negligible.
\item \textbf{Preconditioned correction:} alongside the vanilla update on $\mathbf{W}_N$, the $\mathbf{M}$-path contributes an additional correction term proportional to $(\mathbf{W}_P\mathbf{W}_P^\top)\widehat{\mathbf{G}}$ (Lemma~1), steering part of the update toward directions spanned by $\mathbf{W}_P$.
\item \textbf{Noise filtering:} the correction term maps updates into $\mathrm{col}(\mathbf{W}_P)$, i.e., it removes components in $\mathrm{col}(\mathbf{W}_P)^\perp$.
Moreover, the \emph{orthogonally projected} path $\mathbf{P}\widehat{\mathbf{G}}$ cannot increase the noise energy (Lemma~2); the additional magnitude scaling induced by $\mathbf{W}_P\mathbf{W}_P^\top$ is controlled by $\eta_M$.
\end{itemize}
Under the assumption Eq.\eqref{eq:assump_subspace_snr}, $\mathbf{W}_P$ is trained on signal-rich directions and tends to form a relatively well-conditioned subspace; APEX leverages this subspace to supply stable residual corrections while still allowing $\mathbf{W}_N$ to learn directly when it is informative.

\section{Broader Impact}
The APEX method focuses on efficiency, demonstrating the ability to match the performance of models with more trainable parameters. This effectively reduces the computational cost required to reach target performance and mitigates the environmental impact of model training. Specifically, in instruction tuning, APEX surpassed the performance of full-parameter fine-tuning using only 52\% of the trainable parameters. In continued pre-training, APEX achieved the same perplexity level as conventional training with only approximately 30\% of the pre-training data budget. Moreover, these improvements enable a wider range of research institutions to successfully train larger models within reasonable computational resources, thereby significantly reducing the cost of utilizing large base models.

\section{Algorithm of APEX}
Here, we present the complete algorithm of the proposed method APEX in Algorithm~\ref{alg:apex}.
\begin{algorithm*}
\small
\caption{Advantageous Parameter EXpansion Training (APEX)}
\label{alg:apex}
\begin{algorithmic}[1]
\Require 
    \Statex Model $M$ with trainable parameters $\theta$, which includes $\mathbf{W}_Q,\mathbf{W}_K,\mathbf{W}_V,\mathbf{W}_O,\mathbf{W}_U,\mathbf{W}_D,\mathbf{W}_G$ in each layer
    \Statex Training dataset $\mathcal{D}$
    \Statex Number of stages $T$
    \Statex Proportions $K$
\Ensure Optimized model $M$ with expanded advantageous parameters

\State Initialize $s_{\text{MHA}} \gets 0$, $s_{\text{FFN}} \gets 0$ \Comment{Initialize advantage scores}
\State \textbf{Pre-Assessment Before Training:}
 \For{each batch in $\mathcal{D}_{\text{subset}}$} \Comment{Using a subset of training data}
    \For{each head $h$ in MHA} \Comment{Calculate MHA scores}
        \State $A_{\text{MHA}}^h \gets \text{FrobeniusNorm}(\text{activations})$
        \State Update $s_{\text{MHA}}^h$ with Top-$K$/Min-$K$ counts
    \EndFor
    
    \For{each channel $c$ in FFN} \Comment{Calculate FFN scores}
        \State $A_{\text{FFN}}^c \gets \text{FrobeniusNorm}(\text{activations})$
        \State Update $s_{\text{FFN}}^c$ with Top-$K$/Min-$K$ counts
    \EndFor
\EndFor
\For{stage $t = 0$ \textbf{to} $T-1$}
    \State \textbf{Stage Initialization Phase:}
    \State $d^P_{\text{MHA}} \gets \text{Top-}K (s_{\text{MHA}})$ \Comment{Select parameter sets}
    \State $d^N_{\text{MHA}} \gets \text{Min-}K(s_{\text{MHA}})$
    \State $d^P_{\text{FFN}} \gets \text{Top-}K(s_{\text{FFN}})$
    \State $d^N_{\text{FFN}} \gets \text{Min-}K(s_{\text{FFN}})$
    \State Initialize $\gamma_{\text{MHA}}$ with Monarch matrices ($\mathbf{D}$ initialized randomly, $\mathbf{R}=0$) \Comment{Operator Initialization}
    \State Initialize $\gamma_{\text{FFN}}$ with Monarch matrices ($\mathbf{D}$ initialized randomly, $\mathbf{R}=0$)
    \State
    \State \textbf{Expansion Training Phase:}
    \For{each batch in $\mathcal{D}$} \Comment{Using the training data of the current stage}
        \For{each head $h$ in MHA} \Comment{Calculate MHA scores}
            \State $A_{\text{MHA}}^h \gets \text{FrobeniusNorm}(\text{activations})$
            \State Update $s_{\text{MHA}}^h$ with Top-$K$/Min-$K$ counts
        \EndFor
        
        \For{each channel $c$ in FFN} \Comment{Calculate FFN scores}
            \State $A_{\text{FFN}}^c \gets \text{FrobeniusNorm}(\text{activations})$
            \State Update $s_{\text{FFN}}^c$ with Top-$K$/Min-$K$ counts
        \EndFor
        \State $\text{loss} \gets \text{compute\_loss}(\text{outputs})$ \Comment{Backward Pass}
        \State $\theta \gets \text{optimize}(\text{loss})$ \Comment{Update trainable parameters}
    \EndFor
    \State
    \State \textbf{Operator Fusion Phase:}
    \State $\mathbf{W}_Q[:,d^N_{\text{MHA}}] \gets \gamma_{\text{MHA}}(\mathbf{W}_Q[:,d^P_{\text{MHA}}], \mathbf{W}_Q[:,d^N_{\text{MHA}}])$
    \State $\mathbf{W}_K[:,d^N_{\text{MHA}}] \gets \gamma_{\text{MHA}}(\mathbf{W}_K[:,d^P_{\text{MHA}}], \mathbf{W}_K[:,d^N_{\text{MHA}}])$
    \State $\mathbf{W}_V[:,d^N_{\text{MHA}}] \gets \gamma_{\text{MHA}}(\mathbf{W}_V[:,d^P_{\text{MHA}}], \mathbf{W}_V[:,d^N_{\text{MHA}}])$
    \State $\mathbf{W}_O[d^N_{\text{MHA}},:] \gets \gamma_{\text{MHA}}(\mathbf{W}_O[d^P_{\text{MHA}},:], \mathbf{W}_O[d^N_{\text{MHA}},:])$
    \State $\mathbf{W}_U[:,d^N_{\text{FFN}}] \gets \gamma_{\text{FFN}}(\mathbf{W}_U[:,d^P_{\text{FFN}}], \mathbf{W}_U[:,d^N_{\text{FFN}}])$
    \State $\mathbf{W}_G[:,d^N_{\text{FFN}}] \gets \gamma_{\text{FFN}}(\mathbf{W}_G[:,d^P_{\text{FFN}}], \mathbf{W}_G[:,d^N_{\text{FFN}}])$
    \State $\mathbf{W}_D[d^N_{\text{FFN}},:] \gets \gamma_{\text{FFN}}(\mathbf{W}_D[d^P_{\text{FFN}},:], \mathbf{W}_D[d^N_{\text{FFN}},:])$
\EndFor
\end{algorithmic}
\end{algorithm*}

\end{document}